\documentclass[journal]{IEEEtran}
\usepackage{amsmath,amsfonts}
\usepackage{algpseudocode}
\usepackage{algorithmicx,algorithm}
\usepackage{hyperref}
\usepackage{array}
\usepackage[caption=false,font=normalsize,labelfont=sf,textfont=sf]{subfig}

\usepackage{caption}
\usepackage{textcomp}
\usepackage{stfloats}
\usepackage{url}
\usepackage{verbatim}
\usepackage{graphicx}
\usepackage{cite}

\usepackage[square,sort&compress,numbers]{natbib}
\usepackage{amsmath}
\usepackage{amsfonts} 
\usepackage{amssymb} 
\usepackage{multirow}
\usepackage{threeparttable}
\usepackage{booktabs}

\usepackage{bbm, dsfont}
\usepackage{bbding}
\usepackage{ifsym}
\usepackage{pifont}

\newcommand{\eg}{\textit{e.g.}}

\newcommand{\vmark}{\ding{51}} 
\newcommand{\xmark}{\ding{55}} 

\graphicspath{{figs/}}
\graphicspath{{bio/}}

\usepackage[table]{xcolor}

\definecolor{mygreen}{HTML}{EDF4E5}

\begin{document}
\title{RSRefSeg 2: Decoupling Referring Remote Sensing Image Segmentation with Foundation Models}
\author{
Keyan~Chen,~Chenyang~Liu,~Bowen~Chen,~Jiafan~Zhang,~Zhengxia~Zou,~and~Zhenwei~Shi$^\star$
\\
Beihang University

}

\maketitle

\begin{abstract}

Referring Remote Sensing Image Segmentation (RRSIS) provides a flexible and fine-grained framework for remote sensing scene analysis via vision-language collaborative interpretation. Current approaches predominantly utilize a three-stage pipeline encompassing dual-modal encoding, cross-modal interaction, and pixel decoding. These methods demonstrate significant limitations in managing complex semantic relationships and achieving precise cross-modal alignment, largely due to their coupled processing mechanism that conflates target localization (``\textit{where}") with boundary delineation (``\textit{how}"). This architectural coupling amplifies error propagation under semantic ambiguity while restricting model generalizability and interpretability. To address these issues, we propose \textit{\textbf{RSRefSeg 2}}, a novel decoupling paradigm that reformulates the conventional workflow into a collaborative dual-stage framework: ``\textit{coarse localization followed by fine segmentation}". RSRefSeg 2 integrates CLIP's cross-modal alignment strength with SAM's segmentation generalizability through strategic foundation model collaboration. Specifically, CLIP is employed as the dual-modal encoder to activate target features within its pre-aligned semantic space and generate preliminary localization prompts. To mitigate CLIP's misactivation challenges in multi-entity scenarios described by referring expressions, a cascaded second-order prompter is devised. This prompter enhances precision through implicit reasoning via decomposition of text embeddings into complementary semantic subspaces. These optimized semantic prompts subsequently direct the SAM to generate pixel-level refined masks, thereby completing the semantic transmission pipeline. Furthermore, parameter-efficient tuning strategies are introduced to enhance the domain-specific feature adaptation capabilities of the foundation models for remote sensing applications. Extensive experiments across three benchmark datasets (RefSegRS, RRSIS-D, and RISBench) demonstrate that RSRefSeg 2 surpasses contemporary methods in segmentation accuracy (+$\sim$3\% gIoU improvement) and complex semantic interpretation. The implementation is publicly available at: \url{https://github.com/KyanChen/RSRefSeg2}.

\end{abstract}


\IEEEpeerreviewmaketitle

\section{Introduction}

\IEEEPARstart{R}eferring Remote Sensing Image Segmentation (RRSIS), an advanced domain within vision-language multimodal understanding, aims to achieve precise segmentation of semantic regions or targets in remote sensing imagery guided by natural language textual descriptions \cite{yuan2024rrsis, chen2025rsrefseg, dong2024cross}. By establishing a language-guided visual understanding mechanism, this technology addresses the inherent limitations of conventional remote sensing semantic segmentation frameworks, offering greater flexibility and granularity in image analysis \cite{dong2025diffris}. This paradigm not only enhances user interaction efficiency and analytical precision but also demonstrates significant practical value in applications such as land-use planning, ecological monitoring, and disaster emergency management \cite{liu2024rotated, lei2024exploring, chen2021building, yang2025large, ma2025lscf, liu2022dual, liu2024deriving}.

\begin{figure}[!htb]
\centering
\vspace{-0.4cm}
\resizebox{\linewidth}{!}{
\includegraphics[width=\linewidth]{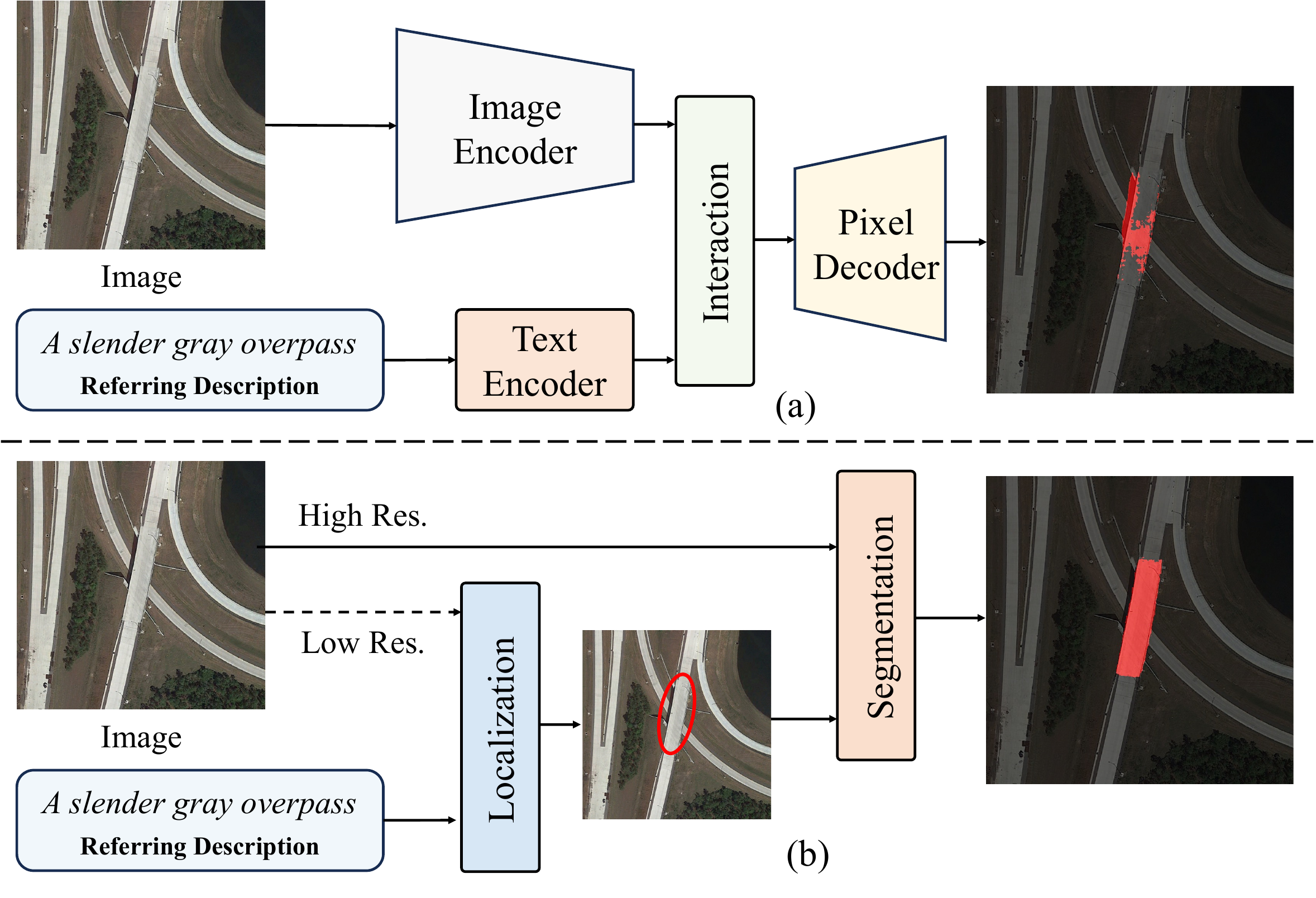}
}
\vspace{-0.4cm}
\caption{
(a) Conventional: three-stage (dual encoders, cross-interaction, pixel decoder). (b) RSRefSeg 2: collaborative coarse localization and fine segmentation.
\label{fig:teaser}}
\vspace{-0.4cm}
\end{figure}

Although groundbreaking progress in multimodal understanding has facilitated preliminary advances in this field, evidenced by the proposal of RRSIS benchmark tasks and the creation of the RefSegRS dataset \cite{yuan2024rrsis, chen2023ovarnet, zou2023object}, persistent limitations remain in existing methodologies \cite{zhang2025referring, ho2024rssep}. Early dual-encoder architectures based on CNN-RNN frameworks achieved rudimentary feature fusion but were constrained by shallow interaction mechanisms, compromising pixel-level segmentation accuracy \cite{yuan2024rrsis}. While subsequent integration of attention mechanisms enhanced semantic space interactions, mainstream approaches remain dependent on pre-trained language models (\eg, BERT) for text encoding and rely on simplistic fusion strategies, such as direct concatenation or unidirectional text-visual interactions \cite{pan2024rethinking, li2025multimodal, liu2025cadformer}. These methods inadequately address the inherent distributional discrepancies between visual and textual features in remote sensing scenarios. Furthermore, the scarcity of paired remote sensing image-text datasets impedes fine-grained semantic alignment, limiting cross-scene generalization \cite{radford2021learning, chen2025dynamicvis}. Current frameworks predominantly adopt three-stage architectures comprising dual encoders, cross-modal interaction modules, and pixel decoders \cite{dong2025diffris, shi2025multimodal, li2025semantic, li2025scale}. However, this unified framework conflates two critical subtasks: target localization (``\textit{where}'') and mask generation (``\textit{how}''), detrimentally affecting both accuracy and interpretability \cite{li2025semantic, li2025aeroreformer}, as demonstrated in Fig.~\ref{fig:teaser} (a). Building on decoupled modeling principles, this work posits that identifying the semantic localization of target objects enables mask generation to proceed independently of textual guidance. This decoupling facilitates specialized optimization for localization and segmentation, improving comprehension accuracy through task-specific refinement. Additionally, it enhances interpretability by differentiating localization errors (indicative of textual comprehension deficiencies) from segmentation errors (attributable to visual representation limitations). Consequently, the proposed strategy establishes a clear diagnostic framework for iterative model refinement, enabling targeted resolution of performance bottlenecks, as shown in Fig.~\ref{fig:teaser} (b).

\begin{figure}[!htb]
\centering
\resizebox{0.98\linewidth}{!}{
\includegraphics[width=\linewidth]{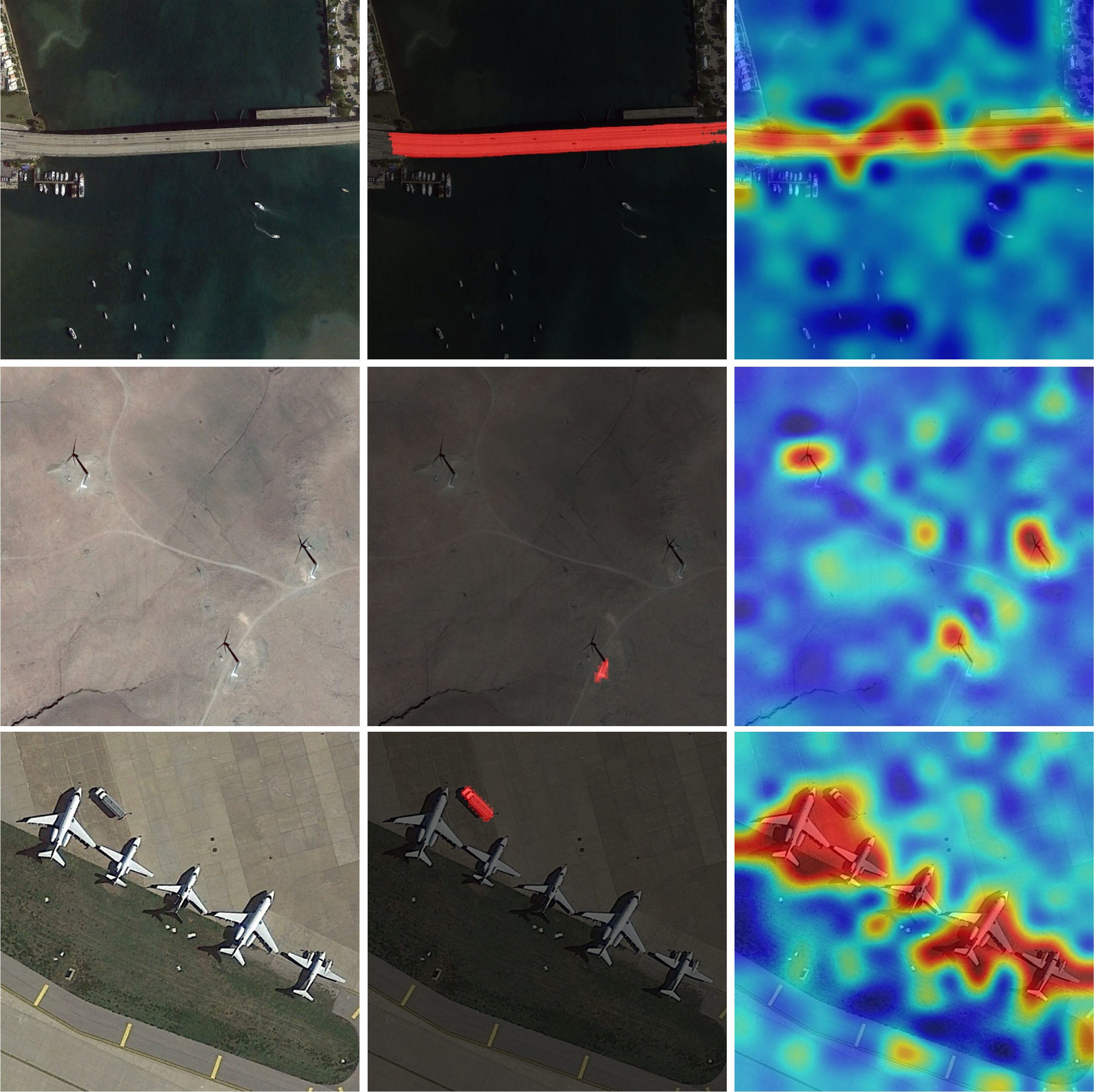}
}
\caption{
Visualization and comparison of the native CLIP's target perception. In simple scenarios, CLIP accurately activates the target region. However, when multiple objects of the same category are present (second row), or when the main entities do not match the referred target (third row), CLIP produces multiple or incorrect activation regions. The referring texts from top to bottom are: ``The orange and gray large bridge in the middle," ``The windmill at the bottom," and ``The vehicle is on the right of the airplane on the upper left." \label{fig:fig_clip_misactivation}
}
\vspace{-0.5cm}
\end{figure}

This paper introduces RSRefSeg 2, a decoupled framework for referring remote sensing image segmentation that decomposes the semantic-driven task into two sequential stages: coarse-grained object localization and fine-grained mask generation. As depicted in Fig.~\ref{fig:fig_model}, the framework leverages the cross-modal alignment capabilities of CLIP \cite{radford2021learning} and the high-generalization segmentation capacity of SAM \cite{kirillov2023segment} to establish a robust open-world remote sensing image understanding system \cite{liu2024change}. A pretrained CLIP model is employed as a dual-modality encoder, where a prompt generator activates text-relevant visual features within the pre-aligned text-visual semantic space to produce SAM-compatible localization prompts. However, the native CLIP model’s sentence-level alignment characteristics risk target misactivation when processing multi-entity text descriptions (Fig. \ref{fig:fig_clip_misactivation}). To address this limitation, a cascaded second-order referring prompting mechanism is introduced (Fig. \ref{fig:fig_model}, bottom right). This mechanism decomposes original text embeddings into complementary semantic subspaces via projection: the first subspace identifies potential target regions through cross-modal feature interaction, while the second refines localization prompts using activated visual feature maps. Optimized under orthogonal constraints and task-specific objectives, the mechanism facilitates implicit cascaded reasoning, enhancing semantic accuracy in multi-entity scenarios while preserving end-to-end trainability for potential reinforcement learning extensions. Additionally, low-rank adaptation is incorporated to address domain shift challenges of vision foundation models in remote sensing, enabling parameter-efficient tuning that balances domain-specific adaptation with general knowledge retention.

The primary contributions of this work are threefold:

i) A dual-stage decoupled framework, RSRefSeg 2, is introduced to enhance decision interpretability by decomposing referring segmentation into sequential ``\textit{slack localization}" and ``\textit{refined segmentation}" subtasks. By leveraging cross-modal priors from vision foundation models, the framework achieves systematic improvements in open-world semantic understanding and cross-scenario generalization.

ii) A cascaded second-order prompting mechanism is designed to resolve CLIP’s semantic ambiguity in multi-entity contexts. Through the decomposition of text embeddings and enabling implicit cascaded reasoning, this mechanism establishes an efficient semantic transmission bridge between CLIP and SAM, effectively mitigating target misactivation in complex scenes.

iii) Extensive experiments across three remote sensing benchmarks (RefSegRS, RRSIS-D, RISBench) demonstrate the framework’s superiority over existing methods in segmentation accuracy and semantic comprehension. Ablation studies further corroborate the architectural coherence and module-level synergies.

\vspace{-0.2cm}
\section{Related Works}

\subsection{Referring Remote Sensing Image Segmentation}

RRSIS has emerged as a pivotal advancement in visual-language multimodal understanding, enabling precise segmentation of semantically relevant regions or targets in remote sensing imagery through natural language descriptions \cite{chen2025rsrefseg, shi2025multimodal}. This capability provides novel technical pathways for interactive geospatial analysis. However, two primary challenges persist: i) inherent characteristics of remote sensing images, including significant scale variations, complex scene compositions, and low foreground-background contrast, complicate precise object boundary delineation \cite{chen2022resolution, liu2025text2earth}; and ii) accurate cross-modal alignment is required to interpret open geospatial relationships at fine-grained semantic levels \cite{chen2024rsmamba, li2021geographical, li2025segearth}.

Early methods predominantly utilized dual-stream architectures combining CNNs and RNNs to achieve pixel-level predictions via shallow feature fusion \cite{yang2022lavt}. Subsequent advancements in multimodal learning established a three-stage \textit{``representation-fusion-segmentation"} paradigm, where modality-specific features extracted by pretrained language models and visual encoders undergo multimodal fusion to generate masks \cite{pan2025mixed, lei2024exploring, liu2024rotated}. Research has since focused on cross-modal interaction mechanisms, diverging into three categories: i) Simple feature concatenation approaches directly align textual and visual features. For example, LAVT integrates BERT-derived text features through visual Transformers \cite{yang2022lavt}, while LGCE enhances this with cross-layer adaptive fusion \cite{yuan2024rrsis}. ii) Text-guided visual representation methods modulate visual features using referring information. RMSIN employs intra-scale and cross-scale interaction modules to capture fine-grained spatial details \cite{liu2024rotated}, whereas FIANet decouples entity semantics and spatial relationships via semantic decomposition networks \cite{lei2024exploring}. iii) Bidirectional cross-modal interaction frameworks refine language representations using visual feedback. Architectures such as BTDNet \cite{zhang2025referring} and CroBIM \cite{dong2024cross} improve feature discriminability through bidirectional attention, while SBANet introduces scale-aware cross-modal queries for efficient feature selection \cite{li2025scale}.  

Notably, unlike natural imagery where models like CLIP achieve robust vision-language alignment, the substantial modality gap in remote sensing restricts existing methods \cite{li2022geographical, chen2024time}. Limited training data further impedes fine-grained semantic alignment, resulting in suboptimal feature generalization across diverse geospatial contexts \cite{liu2024rscama, chen2023continuous}. To mitigate these limitations, RSRefSeg 2 introduces a decoupled \textit{``localization-segmentation"} architecture based on pre-aligned vision-language models for knowledge transfer, reducing multi-stage optimization complexity while enhancing interpretability compared to traditional three-stage frameworks.

\vspace{-0.2cm}
\subsection{Foundation Models}

The rapid advancement of foundation models has catalyzed a paradigm shift in artificial intelligence, characterized primarily by exceptional task-processing capabilities and cross-domain generalization properties \cite{chen2025seg, bai2023qwen}. Current foundation model systems predominantly consist of text-driven large language models, while vision foundation models, though less prevalent, exhibit significant technological progress \cite{radford2021learning, kirillov2023segment}. CLIP demonstrates powerful zero-shot recognition capabilities by leveraging large-scale image-text contrastive learning \cite{radford2021learning}. The Diffusion model series surpasses traditional GANs in image generation quality through a progressive denoising architecture \cite{liu2025text2earth, rombach2022high}. SAM addresses generalization challenges in universal image segmentation via an automated annotation engine \cite{kirillov2023segment, ravi2024sam}. Concurrently, remote sensing foundation models have undergone accelerated iteration \cite{chen2024rsprompter, liu2024remoteclip, guo2024skysense, wang2024mtp, li2025agrifm}: the RingMo series achieves superior multi-task performance via billion-parameter designs \cite{sun2022ringmo, bi2025ringmoe}; SpectralGPT and HyperSigma extend processing capabilities to multispectral and hyperspectral data, respectively \cite{hong2024spectralgpt, wang2025hypersigma}; and DynamicVis innovatively balances precision and efficiency in multi-task optimization through a token-reduction state-space architecture \cite{chen2025dynamicvis}.

Pre-trained model fine-tuning offers dual advantages: it reduces sample dependency requirements in high-precision scenarios while simultaneously enhancing cross-scenario generalization \cite{chen2024rsprompter, hu2025diffusion}. Representative applications exemplifying these benefits include AnyChange for arbitrary change detection \cite{zheng2024segment}, CrossEarth as the first domain-generalized semantic segmentation model in remote sensing \cite{gong2024crossearth}, and SegEarth-R1, which achieves semantic-guided segmentation through hierarchical visual-language encoder collaboration \cite{li2025segearth}.

The proposed RSRefSeg 2 framework advances foundation model fine-tuning by integrating knowledge from general vision models to enhance referring segmentation performance under open-vocabulary text descriptions and open-set remote sensing scenarios. Task-specific optimization is achieved through knowledge transfer mechanisms operating under constraints of limited data and computational resources. This approach offers practical solutions to real-world challenges such as sample scarcity and computational bottlenecks.

\vspace{-0.3cm}
\subsection{Parameter-Efficient Fine-Tuning}

In recent years, large-scale foundation models have surpassed traditional machine learning paradigms, which predominantly relied on fully supervised learning. Vision foundation models, including CLIP, DINO, BLIP, and SAM \cite{radford2021learning, zhang2022dino, li2022blip, kirillov2023segment}, demonstrate exceptional transferability and generalization capabilities across diverse downstream tasks. Prompt learning has emerged within this context as a parameter-efficient adaptation strategy \cite{zhou2022learning}. It bridges the semantic discrepancy between pre-trained models and task-specific requirements through task-aligned prompt engineering, requiring minimal or no fine-tuning of the base model's parameters. Originally derived from natural language processing, prompt learning operates by eliciting latent knowledge representations through strategically designed input cues, with the pre-trained parameters remaining fixed. In visual domains, this manifests through two principal methodologies: \textit{text prompt optimization} for vision-language models and \textit{visual prompt optimization} for pure vision architectures \cite{ding2021openprompt, khattak2023maple, zhou2022conditional, lei2024prompt}.

Parameter-efficient fine-tuning (PEFT) methodologies have gained prominence through architectural innovations such as Adapters, Low-Rank Adaptation (LoRA), and Prefix Tuning. These approaches enhance few-shot adaptation and task specialization by integrating lightweight modules while keeping backbone parameters frozen \cite{han2024parameter, xin2024parameter}. For instance, CoOp and VPT demonstrate significant improvements in few-shot classification \cite{zhou2022learning, jia2022visual}, while RSPrompter equips the SAM with automated instance segmentation capabilities for remote sensing imagery via prompt-guided adaptation \cite{chen2024rsprompter}.

This work introduces a decoupled architecture for referring remote sensing image segmentation with foundation models. Cross-domain knowledge transfer is achieved through the collaborative adaptation of CLIP and SAM architectures; CLIP processes semantic referring information, while SAM generates segmentation masks. Furthermore, a novel cascaded second-order prompting mechanism is proposed to propagate contextual information between the localization and segmentation phases, enabling joint optimization of both foundation models. This co-design significantly reduces trainable parameters compared to full fine-tuning while improving segmentation accuracy in complex scenarios characterized by occlusions, ambiguous boundaries, and extreme scale variations.

\section{Methodology}

\subsection{Overview}

\begin{figure*}[!tb]
\centering
\resizebox{0.99\linewidth}{!}{
\includegraphics[width=\linewidth]{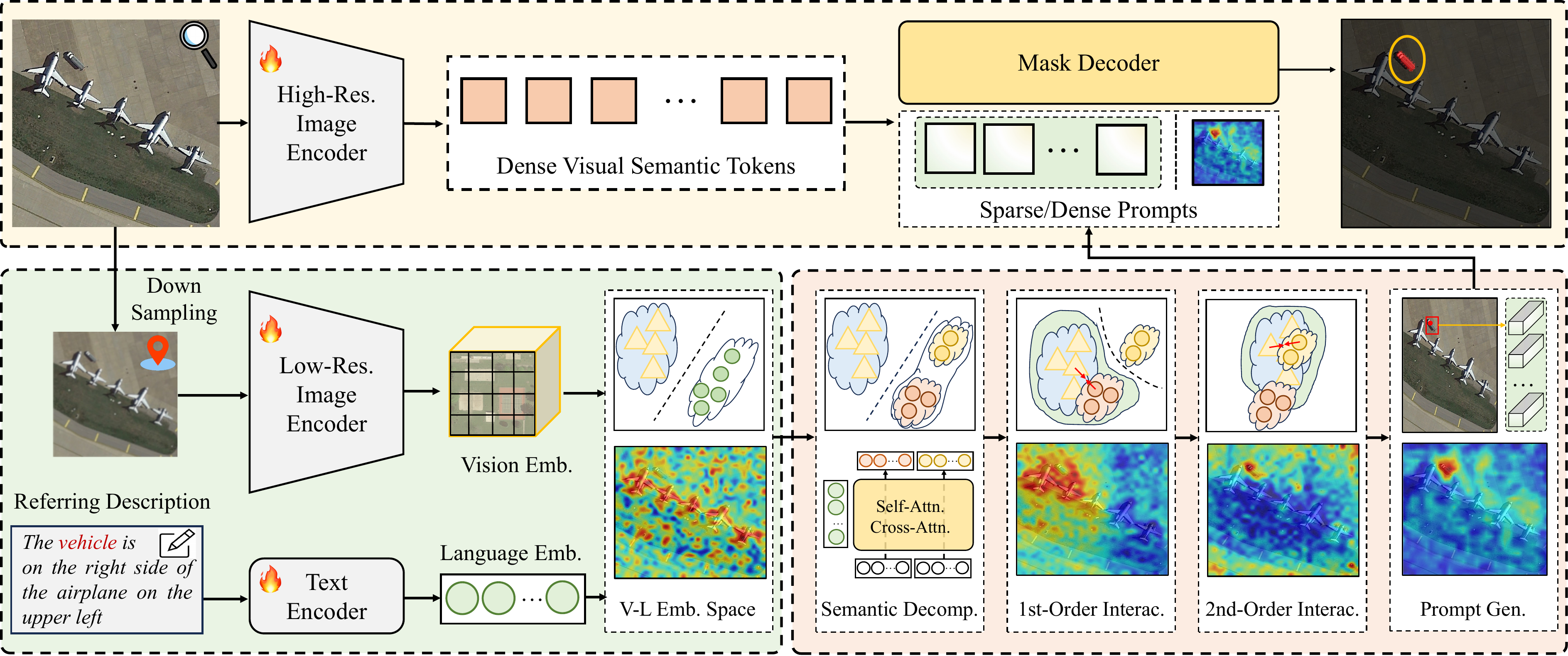}
}
\caption{
The architecture of RSRefSeg 2 consists of three main components: the dual-modality aligned semantic feature encoder ( \raisebox{1.7pt}{\fcolorbox{black}[HTML]{EDF4E5}{\rule[0pt]{0pt}{2pt}\hspace{6pt}}} ), the cascaded second-order referring prompter ( \raisebox{1.7pt}{\fcolorbox{black}[HTML]{FAEEE3}{\rule[0pt]{0pt}{2pt}\hspace{6pt}}} ), and the prompt-guided refinement mask generator ( \raisebox{1.7pt}{\fcolorbox{black}[HTML]{FEF8E7}{\rule[0pt]{0pt}{2pt}\hspace{6pt}}} ). \label{fig:fig_model}}
\vspace{-0.5cm}
\end{figure*}

This paper introduces RSRefSeg 2, a framework that strategically decomposes the complex segmentation process into two sequential stages: coarse-grained object localization followed by fine-grained mask generation. By leveraging CLIP's cross-modal alignment capabilities and SAM's high-generalization segmentation strengths, the framework advances semantic understanding for open-vocabulary referring scenarios in remote sensing imagery. As illustrated in Fig. \ref{fig:fig_model}, the architecture comprises three principal components: a dual-modality aligned semantic feature encoder, a cascaded second-order referring prompter, and a prompt-guided refinement mask generator.

Within the dual-modality encoder, textual referring expressions and visual features derived from low-resolution images are simultaneously processed within a pre-aligned semantic space, establishing unified cross-modal coarse-grained representations. Subsequently, the cascaded second-order referring prompter generates spatially activated localization information through semantic deconstruction of text embeddings and implicit multimodal feature interactions, which are encoded as spatially-aware prompts. Finally, the refinement mask generator synthesizes pixel-level segmentation masks by integrating spatial details from high-resolution original images with prompt-derived contextual information. The proposed processing pipeline is formally defined as:
\begin{align}
\begin{split}
\mathcal{F} &= \Phi_\text{enc}(\mathcal{I}_1, \mathcal{T}) \\
\mathcal{P} &= \Phi_\text{prompter}(\mathcal{F}) \\
\mathcal{\hat{M}} &= \Phi_\text{dec}(\mathcal{I}_2, \mathcal{P}) \\
\end{split}
\end{align}
where $\mathcal{I}_1 \in \mathbb{R}^{H_1 \times W_1 \times 3}$ and $\mathcal{I}_2 \in \mathbb{R}^{H_2 \times W_2 \times 3}$ denote the downsampled low-resolution and original high-resolution images, respectively, while $\mathcal{T} \in \mathbb{R}^{N}$ represents the textual referring description. The dual-modality encoder $\Phi_{\text{enc}}(\cdot)$ separately encodes image and text features to produce $\mathcal{F} = \{v, t\}$, where $v \in \mathbb{R}^{h_1 \times w_1 \times d_1}$ corresponds to the visual feature map and $t \in \mathbb{R}^{(L+1) \times d_1}$ encapsulates textual features comprising $L$ semantic tokens and one global \texttt{<EOS>} token.

Hybrid prompts $\mathcal{P}$ are subsequently generated by $\Phi_\text{prompter}(\cdot)$ through integration of pre-aligned visual–text features, combining sparse positional embeddings (point/box coordinates) with a dense segmentation probability map. These prompts are leveraged by the decoder $\Phi_\text{dec}(\cdot)$ to synthesize refined binary masks $\mathcal{\hat{M}} \in \mathbb{R}^{H_2 \times W_2}$, achieved through fusion of high-resolution spatial information from $\mathcal{I}_2$ with prompt-guided contextual cues.

This hierarchical architecture reduces cross-modal alignment complexity via feature decoupling while preserving fine spatial details through residual high-resolution processing. The framework capitalizes on SAM's prompt adaptability to maintain segmentation accuracy despite relaxed spatial prompt precision under computational constraints ($H_1 < H_2$, $W_1 < W_2$). By decoupling semantic interpretation from mask refinement, the two-stage design effectively balances generalization requirements for open-vocabulary understanding with precision demands for pixel-level segmentation.

\vspace{-0.4cm}
\subsection{Dual-modality Aligned Semantic Feature Encoder} \label{sec:dual-encoder}

RSRefSeg 2 utilizes the text-visual joint semantic space of the CLIP pretrained model to enhance cross-scene generalization capabilities for open-set textual referring. This is achieved by extracting pre-aligned textual semantics and visual features. CLIP provides robust encoding of free-form text and visual inputs while establishing cross-modal semantic alignment \cite{radford2021learning}. However, two critical challenges limit performance in referring remote sensing image segmentation tasks. First, the complex characteristics of remote sensing images, including intricate spatial relationships and attribute descriptions within referring instructions, significantly deviate from natural language description paradigms \cite{li2023rs}. Second, coarse-grained image-sentence alignment often results in imprecise target localization, manifested as either misalignment across multiple targets or incomplete activation \cite{zhong2022regionclip}. To address these limitations, a low-rank adaptation-based domain transfer method is introduced, optimizing cross-modal feature alignment as follows:  
\begin{align}
\begin{split}
t&= \Phi_\text{enc-t}(\mathcal{T}, W_t, W^\star_t) \\
v&= \Phi_\text{enc-v}(\mathcal{I}_1, W_v, W^\star_v) \\
W^\star &= AB^T \label{eq:feat_encoder}
\end{split}
\end{align}
where $\Phi_\text{enc-t}$ and $\Phi_\text{enc-v}$ denote CLIP's text and visual encoders, respectively. $W \in \mathbb{R}^{d \times d}$ represents the frozen original parameter matrix, while $W^\star$ corresponds to trainable low-rank parameters composed of $A \in \mathbb{R}^{d \times r}$ and $B \in \mathbb{R}^{d \times r}$, with $r \ll d$ enforcing a rank constraint \cite{hu2022lora}. RSRefSeg 2 employs a decoupled design, shifting focus from CLIP's pre-established fine-grained alignment capabilities in the remote sensing domain to address target slack localization. Final segmentation accuracy is then achieved through a dedicated downstream module optimized for fine-grained region delineation.

To address the demand for dense features in referring segmentation, this study enhances the representational capacity of the original CLIP. Whereas naive CLIP produces sparse image and text representations, the visual encoder is modified by removing its pooling layer, thereby preserving spatial resolution to generate a dense visual feature map $v \in \mathbb{R}^{h_1 \times w_1 \times d_1}$. For textual processing, word-level features $t_{\text{word}} \in \mathbb{R}^{L \times d_1}$ are derived by aggregating semantic embeddings from each token in the final hidden states of the text encoder. These features explicitly encode fine-grained details such as category, attribute, and spatial cues from the referring expression. Additionally, the output feature at the End-of-Sequence (EOS) token position is extracted as the sentence-level embedding $t_{\text{sent}} \in \mathbb{R}^{1 \times d_1}$, which encapsulates the global semantics of the entire text. The final fused text feature $t$ is obtained through concatenation: $t = [t_{\text{word}}, t_{\text{sent}}]$. This approach achieves cohesive integration of local and global textual semantics with spatially preserved visual features, optimizing the model for dense prediction.

\vspace{-0.4cm}
\subsection{Cascaded Second-order Referring Prompter} \label{sec:prompter}

The RSRefSeg 2 prompter activates visual features relevant to the textual description within a pre-aligned text-visual semantic space, subsequently generating coarse localization prompts for the SAM. However, experimental results indicate that in scenes characterized by complex semantics, features activated by the prompter may correspond to non-target spatial locations. For example, comprehending intricate semantic expressions such as ``The vehicle is on the right side of the airplane above" necessitates a rigorous reasoning chain: first locating the airplane, then identifying its upper image position, and finally determining the vehicle positioned to its right. By contrast, the native CLIP primarily aligns the entire image globally with the complete sentence, emphasizing core semantic entities (\eg, ``airplane") while frequently neglecting the constraining influences of attribute descriptors (\eg, ``above") and relative positional relationships (\eg, ``on the right of") on target selection. This tendency often yields multiple or erroneous target activations.

To address these limitations, this paper proposes a cascaded second-order referring prompting mechanism. The approach initially decomposes original text embeddings into complementary semantic subspaces. Iterative text-visual feature interactions then implicitly locate candidate semantic regions and further activate specific targets. Due to the absence of explicit reasoning step information, this process is achieved through joint optimization of an orthogonal constraint on the complementary semantic subspaces and the segmentation task objective, termed implicit cascaded reasoning. The design additionally establishes an extensible interface for potential future reinforcement learning-based optimization. Specifically, the prompter comprises four modules: Referring Semantic Decomposition, First-order Text-Visual Interaction, Second-order Text-Visual Interaction, and Sparse-Dense Prompt Generation as follows:
\begin{align}
\begin{split}
t_1, t_2 &= \Phi_\text{decomp}(t_{\text{word}}) \\
t^\prime_1, v^\prime_1 &= \Phi_\text{inter-act}(t_1, v) \\
t^\prime_2, v^\prime_2 &= \Phi_\text{inter-act}(t_2, v^\prime_1) \\
\mathcal{P} &= \Phi_\text{prompt}(t^\prime_2, v^\prime_2, t_{\text{sent}}) \label{eq:prompter}
\end{split}
\end{align}
where $t_{\text{word}} \in \mathbb{R}^{L \times d_1}$ represents word-level text embeddings; $v \in \mathbb{R}^{h_1 \times w_1 \times d_1}$ denotes the input visual features; and $t_{\text{sent}} \in \mathbb{R}^{1 \times d_1}$ corresponds to sentence-level text embeddings. The text embeddings for complementary semantic subspaces are denoted by $t_1, t_2 \in \mathbb{R}^{n_t \times d_1}$. Subsequently, $t^\prime_1 \in \mathbb{R}^{n_t \times d_1}$, $v^\prime_1 \in \mathbb{R}^{h_1 \times w_1 \times d_1}$ and $t^\prime_2 \in \mathbb{R}^{n_t \times d_1}$, $v^\prime_2 \in \mathbb{R}^{h_1 \times w_1 \times d_1}$ represent the text embeddings and visual features activated by the first-order and second-order interactions, respectively. The generated prompt information $\mathcal{P}$ comprises sparse prompts $p_{\text{sparse}} \in \mathbb{R}^{n_p \times d_2}$ and dense prompts $p_{\text{dense}} \in \mathbb{R}^{\frac{H_2}{4} \times \frac{W_2}{4}}$. $\Phi_\text{decomp}$, $\Phi_\text{inter-act}$, and $\Phi_\text{prompt}$ denote the Referring Semantic Decomposition, Text-Visual Interaction Activation, and Prompt Generation modules, respectively.

\noindent \textbf{Referring Semantic Decomposition}: The core concept of Referring Semantic Decomposition involves decoupling the original referring expression into two distinct semantic subspaces. This decomposition process, denoted ${\Phi_\text{decomp}}$, is implemented via a cross-attention mechanism between predefined learnable embedding vectors representing the target subspaces and the original referring embedding. To foster information exchange between the resultant subspace embeddings, thereby enhancing joint optimization and feature representation learning, a bidirectional attention interaction module is introduced. Within this module, the embedding vectors of each subspace alternately serve as queries for computing attention across the other subspace, as follows:
\begin{align}
\begin{split}
t_1^q &= \Phi_\text{ln}(\Phi_\text{self-attn}(t_1^q)) \\
t_1 &= \Phi_\text{ln}(\Phi_\text{mlp}(\Phi_\text{ln}(\Phi_\text{cross-attn}(t_1^q, t_{\text{word}})))) \\
t_1 &= \Phi_\text{ln}(\Phi_\text{mlp}(\Phi_\text{ln}(\Phi_\text{cross-attn}(t_1, t_2)))) \label{eq:decomposition}
\end{split}
\end{align}
The above equations model the flow for the first subspace semantic embedding ${t_1 \in \mathbb{R}^{n_t \times d_1}}$. Here, ${t_1^q}$ denotes a preset learnable subspace text embedding vector, which may be initialized through either zero initialization or the pooled representation of the original referring embedding. The modeling process for the second subspace embedding ${t_2 \in \mathbb{R}^{n_t \times d_1}}$ follows the same structure. In the equations, ${\Phi_\text{self-attn}(A)}$ represents a self-attention computation where the query, key, and value are all derived from $A$; similarly, ${\Phi_\text{ln}}$ and ${\Phi_\text{mlp}}$ correspond to layer normalization and multi-layer perceptron projection operations, respectively. The term ${\Phi_\text{cross-attn}(A, B)}$ signifies a cross-attention computation using $A$ as the query and $B$ as both the key and value. For clarity, parameter differences distinguishing these operators are not explicitly annotated, though each represents an independently initialized instance. Collectively, these components define one Referring Semantic Decomposition block. To enhance decomposition capability, ${N_\text{decomp}}$ functionally identical blocks are stacked.

\noindent \textbf{Text-Visual Interaction}: The original text embeddings are decomposed into two semantic subspaces, which are modeled through a cascaded interaction mechanism to enhance visual information expression relevant to referring descriptions. To extract text-aligned visual features, an attention feature activation module, ${\Phi_\text{inter-act}}$, is introduced, utilizing visual features as queries and the semantic embeddings from both decomposed subspaces as keys and values, respectively. The interaction is defined as follows:
\begin{align}
\begin{split}
t^\prime_1 &= \Phi_\text{ln}(\Phi_\text{self-attn}(t_1)) \\
v^\prime_1 &= \Phi_\text{ln}(\Phi_\text{mlp}(\Phi_\text{ln}(\Phi_\text{cross-attn}(v, t^\prime_1))))) \label{eq:Interaction}
\end{split}
\end{align}
This describes interactive activation between the first text semantic subspace embedding $t_1$ and original visual features $v$. The analogous process for the second subspace $t_2$ is omitted for brevity, where input visual features are replaced by the first-stage output $v^\prime_1 \in \mathbb{R}^{h_1 \times w_1 \times d_1}$, establishing a cascaded structure. By stacking $N_\text{inter-act}$ layers of this module, a second-order text-visual semantic interaction framework is formed, yielding refined visual feature representations that precisely localize target objects.

\noindent \textbf{Sparse-Dense Prompt Generation}: 
SAM relies entirely on prompts for segmentation, categorized as sparse prompts (\eg, points, bounding boxes) or dense prompts (\eg, coarse masks) \cite{kirillov2023segment}. However, SAM lacks inherent semantic understanding and cannot generate segmentations automatically based solely on referring text \cite{chen2024rsprompter}. Motivated by this limitation, our approach proposes that the localization stage should exclusively generate coarse positional information aligned with the referring text; consequently, the segmentation stage entirely depends on this positional information. To activate SAM's capability for precise target delineation, corresponding prompt information is generated based on target visual features derived from cross-modal interaction. To fully leverage SAM’s potential, this paper investigates the incorporation of both sparse and dense prompting ($\Phi_\text{prompt}$). Notably, instead of directly generating point coordinates, their corresponding embedding representations are generated to optimize network efficiency. Specifically, the sparse prompt $p_{\text{sparse}} \in \mathbb{R}^{n_p \times d_2}$, which encodes positional information, is generated from the cascaded-interacted visual feature $v^\prime_2$ and text feature $t^\prime_2$:
\begin{align}
\begin{split}
p_{\text{sparse}}^q &= \Phi_\text{ln}(\Phi_\text{self-attn}(p_{\text{sparse}}^q)) 
\\
p_{\text{sparse}} &= \Phi_\text{ln}(\Phi_\text{mlp}(\Phi_\text{ln}(\Phi_\text{cross-attn}(p_{\text{sparse}}^q, t^\prime_2)))) 
\\
p_{\text{sparse}} &= \Phi_\text{ln}(\Phi_\text{mlp}(\Phi_\text{ln}(\Phi_\text{cross-attn}(p_{\text{sparse}}, v^\prime_2))))  \label{eq:prompt-sparse}
\end{split}
\end{align}
where a randomly initialized query vector $p_{\text{sparse}}^q$ sequentially undergoes self-attention, cross-attention with text feature $t^\prime_2$, and cross-attention with visual feature $v^\prime_2$. This processing module is iterated $N_\text{p-gen}$ times to progressively enhance prompt semantics. The aggregated output is subsequently transformed into sparse positional encoding prompts compatible with SAM's requirements. The dense prompt $p_{\text{dense}} \in \mathbb{R}^{\frac{H_2}{4} \times \frac{W_2}{4}}$ is derived from an activation probability map generated through multimodal interaction between visual feature $v^\prime_2$ and whole-sentence text semantic $t_{\text{sent}}$:
\begin{align}
\begin{split} 
F_{\text{dense}} &= \Phi_\text{norm}(\Phi_\text{up}(\Phi_\text{conv}(v^\prime_2)))
\\
t_{\text{sent}} &= \Phi_\text{norm}(t_{\text{sent}}) 
\\
p_{\text{dense}} &= \Phi_\text{einsum}(F_{\text{dense}}, t_{\text{sent}}) \label{eq:prompt-dense}
\end{split}
\end{align}
where $\Phi_\text{conv}$ applies a $3 \times 3$ convolutional kernel with stride 1 and BatchNorm-ReLU layers to smooth visual features. $\Phi_\text{up}$ performs bilinear interpolation to upsample low-resolution CLIP encoder feature maps ($h_1 \times w_1$) to dimensions ($\frac{H_2}{4} \times \frac{W_2}{4}$) compatible with SAM's dense prompt input requirements. $\Phi_\text{norm}$ implements L2-normalization across features. $\Phi_\text{einsum}$ computes a dot product between upsampled feature map $F_{\text{dense}} \in \mathbb{R}^{\frac{H_2}{4} \times \frac{W_2}{4} \times d_1}$ and whole-sentence text embedding $t_{\text{sent}} \in \mathbb{R}^{1 \times d_1}$, thereby generating the activation probability map $p_{\text{dense}}$. The jointly generated sparse and dense prompts are subsequently input to SAM to segment the referred target.

\vspace{-0.4cm}
\subsection{Prompt-guided Refinement Mask Generator}

Although SAM demonstrates exceptional generalization capabilities for segmentation tasks, it lacks inherent task-specific perceptual awareness. RSRefSeg 2 does not directly augment SAM's semantic understanding; instead, it adopts a decoupled paradigm leveraging SAM exclusively as an independent, prompt-based mask generator to fully utilize its general object segmentation ability. Empirical evidence indicates that a strategy of completely freezing SAM's parameters yields suboptimal performance, a limitation largely attributable to the substantial domain discrepancy between natural images and remote sensing data. To mitigate this domain shift with minimal training overhead while maximally preserving SAM's generalizable segmentation knowledge, low-rank trainable parameters are introduced into the ViT backbone, and only the lightweight SAM mask decoder is fine-tuned. This process is formalized as follows:
\begin{align}
\begin{split}
F_{\text{img}} &= \Phi_{\text{img-enc}}(\mathcal{I}_2, W^\star_\text{img}) \\
F_{\text{dense}} &= \Phi_{\text{prompt-enc}}(p_{\text{dense}}) \\
F_{\text{out}} &= \Phi_{\text{concat}}(T_{\text{filter}}, T_{\text{IoU}}, p_{\text{sparse}}) \\
\mathcal{\hat{M}} &= \Phi_{\text{mask-dec}}(F_{\text{img}} + F_{\text{dense}}, F_{\text{out}})
\end{split}
\end{align}
where $\Phi_{\text{img-enc}}$, $\Phi_{\text{prompt-enc}}$, and $\Phi_{\text{mask-dec}}$ denote SAM's image encoder, prompt encoder, and mask decoder, respectively; $W^\star_\text{img}$ represents the learnable low-rank parameters incorporated into the image encoder according to Eq. \ref{eq:feat_encoder}; $\mathcal{I}_2 \in \mathbb{R}^{H_2 \times W_2 \times 3}$ denotes the input high-resolution image; $F_{\text{img}} \in \mathbb{R}^{h_2 \times w_2 \times d_2}$ is the intermediate feature map produced by image encoding; $p_{\text{sparse}} \in \mathbb{R}^{n_p \times d_2}$ (sparse prompt) and $p_{\text{dense}} \in \mathbb{R}^{\frac{H_2}{4} \times \frac{W_2}{4}}$ (coarse mask prompt) are derived from the aforementioned processing flow; $F_{\text{dense}} \in \mathbb{R}^{h_2 \times w_2 \times d_2}$ corresponds to the prompt encoder's output; $T_{\text{filter}} \in \mathbb{R}^{4 \times d_2}$ and $T_{\text{IoU}} \in \mathbb{R}^{1 \times d_2}$ represent predefined learnable tokens for mask filtering and IoU prediction; $\Phi_{\text{concat}}$ denotes token embedding concatenation along the first dimension; and $\mathcal{\hat{M}} \in \mathbb{R}^{4 \times H_2 \times W_2}$ constitutes the multi-mask output. As only a single mask is required, the first mask in $\mathcal{\hat{M}}$ is selected as the final prediction. Technical details are provided in the original SAM paper \cite{kirillov2023segment}.

\vspace{-0.4cm}
\subsection{Loss Function} \label{sec:loss}

The overall loss function comprises three key components: i) $\mathcal{L}_{\text{seg}}$, a supervision loss for the referring segmentation task, integrating binary cross-entropy and Dice loss evaluated on the final predicted mask \cite{chen2025rsrefseg, zhang2025cdmamba}; ii) $\mathcal{L}_{\text{ortho}}$, an orthogonal constraint loss applied to the two semantic subspaces, aiming to maximize their semantic distance through iterative implicit reasoning; and iii) $\mathcal{L}_{\text{align}}$, an alignment loss designed to enhance fine-grained alignment between first-stage visual features and semantic features, which facilitates knowledge transfer from CLIP to the remote sensing referring segmentation domain. The total loss is defined as:
\begin{align}
    \mathcal{L} &= \mathcal{L}_{\text{seg}} + \mathcal{L}_{\text{ortho}} + \mathcal{L}_{\text{align}}
\end{align}

Given the referring mask annotation $\mathcal{M} \in \mathbb{R}^{H_2 \times W_2}$, $\mathcal{L}_{\text{seg}}$ is computed between the predicted mask $\mathcal{\hat{M}}$ and $\mathcal{M}$:
\begin{align}
    \mathcal{L}_{\text{seg}} &= \mathcal{L}_{\text{ce}}(\mathcal{\hat{M}}, \mathcal{M}) + \alpha_{\text{dice}} \mathcal{L}_{\text{dice}}(\mathcal{\hat{M}}, \mathcal{M})
\end{align}

The orthogonal loss $\mathcal{L}_{\text{ortho}}$ is applied to embedding vectors $t_1$ and $t_2$, which are derived from semantic decomposition (Eq. \ref{eq:decomposition}). This loss comprises a normalized average pooling operation, denoted $\Phi_{\text{norm-pool}}$, followed by computation of the squared cosine similarity to minimize feature dependency:
\begin{align}
    \mathcal{L}_{\text{ortho}} &= \alpha_{\text{ortho}} \mathcal{L}_{\text{cos-sim}}\left (\Phi_{\text{norm-pool}}(t_1), \Phi_{\text{norm-pool}}(t_2) \right)
\end{align}
where $\Phi_{\text{norm-pool}}$ is feature-wise normalization applied to average-pooled vectors, reducing tokens to single vectors.

The alignment loss $\mathcal{L}_{\text{align}}$ constrains the first-stage CLIP feature extraction to enhance localization accuracy, thereby improving overall model performance. This loss function integrates three components: i) a dense prompt segmentation constraint ($\mathcal{L}_{\text{dense}}$), ii) a spatial-dimension alignment constraint between CLIP-extracted visual features $v$ and text semantic features $t_\text{sent}$ ($\mathcal{L}_{\text{spat}}$), and iii) a sample-dimension contrastive loss analogous to the original CLIP ($\mathcal{L}_{\text{samp}}$), expressed as:
\begin{align}
\begin{split}
    \mathcal{L}_{\text{align}} &= \alpha_{\text{dense}} \mathcal{L}_{\text{dense}}(p_\text{dense}, \mathcal{M}) + \alpha_{\text{spat}} \mathcal{L}_{\text{spat}}(v, t_\text{sent}, \mathcal{M})  \\ 
    &+ \alpha_{\text{samp}} \mathcal{L}_{\text{samp}}(v, t_\text{sent}, \mathcal{M})
\end{split}
\end{align}
where $\mathcal{L}_{\text{dense}}$ implements a cross-entropy constraint. $\mathcal{L}_{\text{spat}}$ achieves spatial alignment by generating a coarse mask through the Einstein summation of $v$ and $t_\text{sent}$, which is supervised by $\mathcal{M}$ using either cross-entropy (hard supervision) or multi-instance learning (soft supervision) \cite{chen2023ovarnet}. $\mathcal{L}_{\text{samp}}$ enforces sample alignment via a contrastive mechanism, where visual features $v$ undergo annotation-guided mask pooling to form dense representations; these representations and $t_\text{sent}$ constitute positive contrastive pairs. Global contrast is aggregated across all computational nodes to ensure training stability \cite{radford2021learning}. The hyperparameter $\alpha$ balances the contributions of each loss component, the specific impacts of which are analyzed subsequently in ablation experiments.

\vspace{-0.3cm}
\section{Experimental Results and Analyses}

\subsection{Experimental Dataset and Settings}
\label{sec:dataset}

The proposed method was evaluated on three referring remote sensing image segmentation benchmarks: RefSegRS \cite{yuan2024rrsis}, RRSIS-D \cite{liu2024rotated}, and RISBench \cite{dong2024cross}. A statistical summary of these benchmarks is provided in Tab. \ref{tab:dataset_comparison}.

\begin{table}[!h]
\vspace{-0.1cm}
\centering
\caption{Benchmarks for referring remote sensing image segmentation.} \label{tab:dataset_comparison}
\vspace{-0.2cm}
\resizebox{1\linewidth}{!}{
\begin{tabular}{l | *{6}{c}}
\toprule
\textbf{Dataset} & \textbf{Source} & \textbf{Resolution} & \textbf{Size} & \textbf{Triplets}  & \textbf{Categories} & \textbf{Attributes} \\
\midrule
RefSegRS~\cite{yuan2024rrsis} & SkyScapes & 0.13 m & 512$^2$ & 4,420 & 14 & 5 \\
RRSIS-D~\cite{liu2024rotated} & RSVGD & 0.5-30 m & 800$^2$ & 17,402 & 20 & 7 \\
RISBench~\cite{dong2024cross} & VRSBench & 0.1-30 m & 512$^2$ & 52,472  & 26 & 8\\
\bottomrule
\end{tabular}}
\vspace{-0.3cm}
\end{table}

\vspace{3pt}
\noindent \textbf{RefSegRS} \cite{yuan2024rrsis}: 
RefSegRS is constructed by incorporating pixel-level annotations derived from the SkyScapes dataset \cite{azimi2019skyscapes}. It integrates diverse referring expressions paired with corresponding object masks generated through an automated framework. Organized across 285 unique scenes, the dataset contains 4,420 image-description-mask triplets divided into three subsets: a training set (151 scenes, 2,172 triplets), a validation set (31 scenes, 431 triplets), and a test set (103 scenes, 1,817 triplets). Fourteen hierarchical semantic categories are annotated, including roads, buildings, cars, and vans, with each object additionally qualified by five categorical attributes. All images are standardized to a uniform resolution of $512 \times 512$ pixels while preserving the spatial resolution of 0.13 m/pixel.

\vspace{3pt}
\noindent \textbf{RRSIS-D} \cite{liu2024rotated}: 
The dataset is constructed by processing bounding box annotations sourced from the RSVGD dataset \cite{zhan2023rsvg} through SAM to generate instance masks. It encompasses 20 semantic categories, such as aircraft, golf courses, highway service areas, baseball fields, and stadiums. Seven additional attributes supplement these categories to enhance the semantic richness of the referring expressions. Notably, the dataset exhibits substantial scale variation, with numerous targets occupying minimal image areas (\eg, small objects) and others spanning over 400,000 pixels. RRSIS-D contains 17,402 image-description-mask triplets, partitioned into training (12,181 triplets), validation (1,740 triplets), and test (3,481 triplets) subsets. All images are resized to $800 \times 800$ pixels, with spatial resolutions varying between 0.5 and 30 m/pixel.

\vspace{3pt}
\noindent \textbf{RISBench} \cite{dong2024cross}:
RISBench is constructed by integrating remote sensing images, referring expressions, and visual localization bounding boxes sourced from the VRSBench dataset \cite{li2024vrsbench}. This dataset contains 52,472 image-description-mask triplets, partitioned into 26,300 training, 10,013 validation, and 16,158 test samples. All images are standardized to a resolution of $512 \times 512$ pixels and exhibit spatial resolutions ranging between 0.1 and 30 m/pixel to encompass diverse geographical scales and observational granularities. Semantic annotations comprise 26 distinct categories, each augmented by 8 unique attributes to facilitate comprehensive analysis across varying granularities in segmentation tasks. The referring expressions exhibit an average length of 14.31 words, supported by a vocabulary of 4,431 unique words, indicating substantial lexical diversity and syntactic complexity in the textual descriptors.

\vspace{-0.4cm}
\subsection{Evaluation Protocol and Metrics}

Consistent with prior work in referring segmentation \cite{yuan2024rrsis, liu2024rotated}, evaluation employs generalized Intersection over Union (gIoU) and cumulative Intersection over Union (cIoU), defined as:
\begin{align}
\begin{split}
\mathrm{gIoU} &= \frac{1}{N} \sum_{i=1}^{N} \frac{\lvert P_i \cap G_i \rvert}{\lvert P_i \cup G_i \rvert}, \\
\mathrm{cIoU} &= \frac{\sum_{i=1}^{N} \lvert P_i \cap G_i \rvert}{\sum_{i=1}^{N} \lvert P_i \cup G_i \rvert},
\end{split}
\end{align}
where $P_i$ and $G_i$ denote the predicted and ground-truth segmentation masks for the $i$th image, respectively, and $N$ represents the total number of images. Although both metrics are reported, analytical focus is placed on gIoU due to cIoU's inherent bias toward larger target regions. 

To further quantify segmentation quality, precision at IoU thresholds (Pr@$X$) is utilized, which measures the proportion of test images achieving an IoU exceeding a specified threshold $X \in \{0.5, 0.6, \dots, 0.9\}$.

\begin{table*}[!htbp]
\centering
\caption{Performance comparison across various evaluation metrics on the RefSegRS test dataset.} \label{tab:comparisons-RefSegRS}
\vspace{-0.2cm}
\resizebox{0.9\linewidth}{!}{
\begin{tabular}{ c c| *{5}{c} | c c}
\toprule
\textbf{Method} & \textbf{Publication} &\textbf{Pr@0.5} & \textbf{Pr@0.6} & \textbf{Pr@0.7} & \textbf{Pr@0.8} & \textbf{Pr@0.9} & \textbf{cIoU} & \textbf{gIoU} \\ 
\midrule
LSTM-CNN \cite{hu2016segmentation} & ECCV'16 & 15.69 & 10.57 & 5.17 & 1.10 & 0.28 & 53.83 & 24.76 \\
RRN \cite{li2018referring} & CVPR'18 & 30.26 & 23.01 & 14.87 & 7.17 & 0.98 & 65.06 & 41.88 \\
ConvLSTM \cite{li2018referring} & CVPR'18 & 31.21 & 23.39 & 15.30 & 7.59 & 1.10 & 66.12 & 43.34 \\
CMSA \cite{ye2019cross} & CVPR'19 & 28.07 & 20.25 & 12.71 & 5.61 & 0.83 & 64.53 & 41.47 \\
BRINet \cite{hu2020bi} & CVPR'20 & 20.72 & 14.26 & 9.87 & 2.98 & 1.14 & 58.22 & 31.51 \\
LSCM \cite{hui2020linguistic} & ECCV'20 & 31.54 & 20.41 & 9.51 & 5.29 & 0.84 & 61.27 & 35.54 \\
CMPC \cite{huang2020referring}  & CVPR'20 & 32.36 & 14.14 & 6.55 & 1.76 & 0.22 & 55.39 & 40.63 \\
CMPC+ \cite{liu2021cross} & TPAMI'21 & 49.19 & 28.31 & 15.31 & 8.12 & 2.55 & 66.53 & 43.65 \\
CRIS \cite{wang2022cris} & CVPR'22 & 35.77 & 24.11 & 14.36 & 6.38 & 1.21 & 65.87 & 43.26 \\
LAVT \cite{yang2022lavt} & CVPR'22 & 51.84 & 30.27 & 17.34 & 9.52 & 2.09 & 71.86 & 47.40 \\
CARIS \cite{liu2023caris} & ACM MM'23 & 45.40 & 27.19 & 15.08 & 8.87 & 1.98 & 69.74 & 42.66 \\
RIS-DMMI \cite{hu2023beyond} & CVPR'23 & 63.89 & 44.30 & 19.81 & 6.49 & 1.00 & 68.58 & 52.15 \\
CrossVLT \cite{cho2023cross} & TMM'23 & 71.16 & 58.28 & 34.51 & 16.35 & 5.06 & 77.44 & 58.84 \\
EVF-SAM \cite{zhang2024evf} & Arxiv'24 & 35.17 & 22.34 & 9.36 & 2.86 & 0.39 & 55.51 & 36.64 \\
CroBIM \cite{dong2024cross} & Arxiv'24 & 75.89 & 61.42 & 34.07 & 12.99 & 2.75 & 72.33 & 59.77 \\
LGCE \cite{yuan2024rrsis} & TGRS'24 & 73.75 & 61.14 & 39.46 & 16.02 & 5.45 & 76.81 & 59.96 \\
DANet \cite{pan2024rethinking} & ACM MM'24&  76.61 & 64.59 & 42.72 & 18.29 &8.04 &  79.53& 62.14 \\
RMSIN \cite{liu2024rotated} & CVPR'24 & 79.20 & 65.99 & 42.98 & 16.51 & 3.25 & 75.72 & 62.58 \\
FIANet \cite{lei2024exploring} & TGRS'24 & 84.09 & 77.05 & 61.86 & 33.41 & 7.10 & 78.32 & 68.67 \\
SBANet \cite{li2025scale} & Arxiv'25 &77.02 & - & 44.15 & - & 8.97 & 79.86 & 62.73 \\
BTDNet \cite{zhang2025referring} & Arxiv'25 &  83.60  & 75.07 & 62.69 & 34.40 &  9.14  & 80.57 &  67.95 \\
SegEarth-R1 \cite{li2025segearth} & Arxiv'25 &  86.30 & 79.53 &69.57 & 48.87 & 10.73 &79.00 & 72.45 \\
RS2-SAM 2 \cite{rong2025customized} & Arxiv'25 & 84.31 & 79.42 & 70.89 & 55.70 & 21.19 & 80.87 & 73.90 \\
\midrule
RSRefSeg 2 (Ours) & - & \cellcolor{gray!50}\textbf{88.22} & \cellcolor{gray!50}\textbf{82.99} & \cellcolor{gray!50}\textbf{73.97} & \cellcolor{gray!50}\textbf{60.92} & \cellcolor{gray!50}\textbf{34.40} & \cellcolor{gray!50}\textbf{81.24} & \cellcolor{gray!50}\textbf{77.39} \\
\bottomrule
\end{tabular}
}
\vspace{-0.6cm}
\end{table*}

\vspace{-0.4cm}
\subsection{Implementation Details}

RSRefSeg 2 is a decoupled paradigm for referring segmentation by leveraging the complementary strengths of two pre-trained foundation models: the cross-modal alignment capabilities of CLIP and the high generalization capacity for segmentation inherent in SAM \cite{tschannen2025siglip,ravi2024sam}. Unless stated otherwise, the SAM variant employed is `sam2.1-hiera-large'\footnote{https://huggingface.co/facebook/sam2.1-hiera-large}, and the CLIP variant utilized is `siglip2-so400m-patch16-512'\footnote{https://huggingface.co/google/siglip2-so400m-patch16-512}.

\subsubsection{Architecture Details}

To mitigate the domain shift between general-purpose vision models and the remote sensing domain, low-rank fine-tuning parameters are incorporated into their backbone networks with $r = 16$. The lightweight mask decoder from SAM is also integrated within this fine-tuning framework. For the Referring Semantic Decomposition, the quantity of embedding vectors assigned to distinct semantic subspaces is set to $n_t = 3$. Similarly, in the Sparse-Dense Prompt Generation, the number of preset sparse prompt embeddings is configured as $n_p = 9$. Within Referring Semantic Decomposition, the preset subspace embedding vectors ($t^q_1$ and $t^q_2$) are initialized using pooling operations applied to the original referring text representations $t_\text{word}$; learnable positional embeddings are subsequently integrated into these vectors during attention operations. In Sparse-Dense Prompt Generation, the preset sparse prompt query vectors undergo random initialization and similarly incorporate learnable positional embeddings. Unless otherwise specified, the number of blocks is uniformly set to $N_\text{decomp} = N_\text{inter-act} = N_\text{p-gen} = 2$ across all modules. Although SAM is capable of generating diverse output masks, only its first predicted mask is selected as the final output.

\subsubsection{Training Details}

Consistent with the input size requirements of the SAM and CLIP models, training maintained original image dimensions of $H_1=W_1=512$ and $H_2=W_2=1024$. Data augmentation was not employed during training. Fine-tuning was applied exclusively to the newly introduced low-rank parameters, the cascaded prompter parameters, and the SAM decoder head parameters; all other network components remained frozen. The loss balancing coefficients were assigned as follows: $\alpha_\text{dice}=1$, $\alpha_\text{ortho}=\alpha_\text{samp}=0.5$, and $\alpha_\text{dense} = \alpha_\text{spat} =0$. Optimization was performed using AdamW with an initial learning rate of $1\times10^{-4}$, a batch size of 64, and a maximum of 300 epochs. The learning rate was decayed using a Cosine Annealing Scheduler combined with linear warm-up. All experiments were conducted on NVIDIA H800 GPUs. The implementation leveraged the OpenMMLab\footnote{https://github.com/open-mmlab/mmsegmentation} platform and PyTorch framework, with all newly introduced modules trained from scratch. Training efficiency was enhanced using BF16 precision under the DeepSpeed ZeRO Stage 2 distributed framework.

\vspace{-0.3cm}
\subsection{Comparison with the State-of-the-Art}

RSRefSeg 2 is compared with several state-of-the-art referring image segmentation methods, categorized as follows: general vision methods (\eg, CMPC \cite{huang2020referring}, LAVT \cite{yang2022lavt}, CrossVLT \cite{cho2023cross}); remote sensing methods (\eg, RMSIN \cite{liu2024rotated}, FIANet \cite{lei2024exploring}, SBANet \cite{li2025scale}); and foundation model–based methods (\eg, RSRefSeg \cite{chen2025rsrefseg}, RS2‑SAM 2 \cite{rong2025customized}, SegEarth‑R1 \cite{li2025segearth}). Within the result tables, the highest value for each evaluation metric is marked in grey.

\subsubsection{Quantitative Results on the RefSegRS Dataset}

\begin{table*}[!tbp]
\centering
\caption{Performance comparison across various evaluation metrics on the RRSIS-D test dataset.} \label{tab:comparisons-RRSIS-D}
\vspace{-0.2cm}
\resizebox{0.9\linewidth}{!}{
\begin{tabular}{ c c| *{5}{c} | c c}
\toprule
\textbf{Method} & \textbf{Publication} &\textbf{Pr@0.5} & \textbf{Pr@0.6} & \textbf{Pr@0.7} & \textbf{Pr@0.8} & \textbf{Pr@0.9} & \textbf{cIoU} & \textbf{gIoU} \\ 
\midrule
RRN \cite{li2018referring} & CVPR'18 & 51.07 & 42.11 & 32.77 & 21.57 & 6.37 & 66.43 & 45.64 \\
CMSA \cite{ye2019cross} & CVPR'19 & 55.32 & 46.45 & 37.43 & 25.39 & 8.15 & 69.39 & 48.54 \\
CMPC \cite{huang2020referring} & CVPR'20 & 55.83 & 47.40 & 36.94 & 25.45 & 9.19 & 69.22 & 49.24 \\
LSCM \cite{hui2020linguistic} & ECCV'20 & 56.02 & 46.25 & 37.70 & 25.28 & 8.27 & 69.05 & 49.92 \\
BRINet \cite{hu2020bi} & CVPR'20 & 56.90 & 48.77 & 39.12 & 27.03 & 8.73 & 69.88 & 49.65 \\
CMPC+ \cite{liu2021cross} & TPAMI'21 & 57.65 & 47.51 & 36.97 & 24.33 & 7.78 & 68.64 & 50.24 \\
LAVT \cite{yang2022lavt} & CVPR'22 & 69.52 & 63.63 & 53.29 & 41.60 & 24.94 & 77.19 & 61.04 \\
RIS-DMMI \cite{hu2023beyond}& CVPR'23 & 68.74 & 60.96 & 50.33 & 38.38 & 21.63 & 76.20 & 60.12 \\
CrossVLT \cite{cho2023cross}  & TMM'23 & 70.38 & 63.83 & 52.86 & 42.11 & 25.02 & 76.32 & 61.00 \\
LGCE \cite{yuan2024rrsis} & TGRS'24 & 67.65 & 61.53 & 51.45 & 39.62 & 23.33 & 76.34 & 59.37 \\
EVF-SAM \cite{zhang2024evf} & Arxiv'24 & 72.16 & 66.50 & 56.59 & 43.92 & 25.48 & 76.77 & 62.75 \\
FIANet \cite{lei2024exploring} & TGRS'24 & 74.46 & 66.96 & 56.31 & 42.83 & 24.13 & 76.91 & 64.01 \\
RMSIN \cite{liu2024rotated} & CVPR'24 & 74.26 & 67.25 & 55.93 & 42.55 & 24.53 & 77.79 & 64.20 \\
CroBIM \cite{dong2024cross} & Arxiv'24 & 74.58 & 67.57 & 55.59 & 41.63 & 23.56 & 75.99 & 64.46 \\
CADFormer \cite{liu2025cadformer} & JSTARS'25 & 74.20 &67.62& 55.59 &42.37& 23.59 &77.26 &63.77  \\
LSCF \cite{ma2025lscf} & TGRS'25&  74.30 & 67.69  &56.32 &43.08  &25.67  &77.42 & 64.25 \\
RSRefSeg-l \cite{chen2025rsrefseg}	& IGARSS'25 &	74.49&	68.33&	58.73&	48.50&	30.80 &	77.24&	64.67 \\
SBANet \cite{li2025scale}& Arxiv'25  &75.91 & - &57.05 & -  &25.38  &79.22 & 65.52 \\
BTDNet \cite{zhang2025referring} & Arxiv'25 & 75.93 &  69.92&   59.29 &  46.25 &  27.46 & 79.23 & 66.04 \\
SegEarth-R1 \cite{li2025segearth} & Arxiv'25 & 76.96  & - & - & -& -& 78.01  & 66.40 \\
RS2-SAM 2 \cite{rong2025customized} & Arxiv'25 & 77.56 & 72.34 & 61.76 & 47.92 & 29.73 & 78.99 & 66.72 \\
\midrule
RSRefSeg 2 (Ours) & - & \cellcolor{gray!50}\textbf{80.23} & \cellcolor{gray!50}\textbf{75.78} & \cellcolor{gray!50}\textbf{65.41} & \cellcolor{gray!50}\textbf{50.65} & \cellcolor{gray!50}\textbf{31.05} & \cellcolor{gray!50}\textbf{79.45} & \cellcolor{gray!50}\textbf{69.17} \\
\bottomrule
\end{tabular}
}
\vspace{-0.6cm}
\end{table*}

Tab. \ref{tab:comparisons-RefSegRS} presents a comparative performance analysis on the RefSegRS test dataset. Experimental results indicate that RSRefSeg 2 achieves superior performance across all evaluated metrics. Its precision is consistently higher than that of existing methods at varying IoU thresholds, demonstrating enhanced capability for generating accurate segmentation masks. While recent approaches, including RS2-SAM, SegEarth-R1, and BTDNet, have significantly advanced referring image segmentation within remote sensing, RSRefSeg 2 comprehensively outperforms these contemporary solutions. Notably, RSRefSeg 2 achieves significantly higher scores on the composite metrics, attaining 81.24\% cIoU and 77.39\% gIoU. Although incremental improvements in model performance have been observed over time, the innovative design of RSRefSeg 2 advances the state of the art, particularly excelling in precise segmentation tasks requiring higher IoU thresholds.

\subsubsection{Quantitative Results on the RRSIS-D Dataset}

The RRSIS-D dataset exceeds the RefSegRS dataset in scale, encompassing 20 categories and providing approximately a fourfold increase in the number of triplets. As expected, the overall experimental results on RRSIS-D, shown in Tab. \ref{tab:comparisons-RRSIS-D}, demonstrate that the proposed RSRefSeg 2 likewise achieves optimal performance across all evaluation metrics. A declining trend in performance is observed across all methods as thresholds increase; despite this, RSRefSeg 2 exhibits comparatively smaller degradation, indicating superior segmentation accuracy and robustness.

Tab. \ref{tab:comparisons-category-RRSIS-D} displays the performance of compared methods across different categories and reports mean IoU, revealing that RSRefSeg 2 achieves 72.06\% mean IoU, surpassing FIANet (66.46\%). This highlights RSRefSeg 2's significant advantage for referring remote sensing image segmentation. RSRefSeg 2 yields the optimal result in the majority (15 of 20) of categories. It performs particularly well on targets characterized by complex structures yet distinct boundaries (\eg, golf courses, baseball fields, stadiums, chimneys). Furthermore, substantial improvements over alternative methods are demonstrated on challenging small-object categories such as bridges, vehicles, and windmills, indicating its exceptional ability to capture fine-grained details. Minor deficiencies are observed for categories such as dam and harbor, likely due to their ambiguous boundaries which amplify segmentation difficulty.

\begin{table*}[!htbp]
\centering
\caption{
Performance comparison across categories on the RRSIS-D test dataset by class-specific and mean IoU.} \label{tab:comparisons-category-RRSIS-D}
\vspace{-0.2cm}
\resizebox{0.8\linewidth}{!}{
\begin{tabular}{ c | *{4}{c} | c}
\toprule
\textbf{Category} & \textbf{LAVT}  \cite{yang2022lavt} &\textbf{RMSIN} \cite{liu2024rotated} & \textbf{LGCE} \cite{yuan2024rrsis} & \textbf{FIANet} \cite{lei2024exploring} & \textbf{RSRefSeg 2 (Ours)} \\ 
\midrule
airport & 66.44 & 68.08 & 68.11 & 68.66  & \cellcolor{gray!50}\textbf{72.95}\\

golf field & 56.53 & 56.11 & 56.43 & 57.07  & \cellcolor{gray!50}\textbf{79.09}\\
expressway service area & 76.08 & 76.68 & 77.19 & \cellcolor{gray!50}\textbf{77.35}  & 76.49\\
baseball field & 68.56 & 66.93 & 70.93 & 70.44  & \cellcolor{gray!50}\textbf{87.27} \\
stadium & 81.77 & 83.09 & 84.90 & 84.87 &  \cellcolor{gray!50}\textbf{88.73} \\
ground track field & 81.84 & 81.91 & \cellcolor{gray!50}\textbf{82.54} & 82.00  & 79.88\\

storage tank & 71.33 & 73.65 & 73.33 & 76.99 &  \cellcolor{gray!50}\textbf{79.05} \\

basketball court & 70.71 & 72.26 & 74.37 & 74.86 &  \cellcolor{gray!50}\textbf{74.91}\\

chimney & 65.54 & 68.42 & 68.44 & 68.41 &  \cellcolor{gray!50}\textbf{83.97}  \\

tennis court & 74.98 & 76.68 & 75.63 & 78.48 &  \cellcolor{gray!50}\textbf{79.98}\\

overpass & 66.17 & \cellcolor{gray!50}\textbf{70.14} & 67.67 & 70.01 &  66.17\\

train station & 57.02 & 62.67 & 58.19 & 61.30 & \cellcolor{gray!50}\textbf{67.74} \\

ship & 63.47 & 64.64 & 63.48 & 65.96 & \cellcolor{gray!50}\textbf{71.65} \\

expressway toll station & 63.01 & 65.71 & 61.63 & 64.82 &  \cellcolor{gray!50}\textbf{75.03} \\

dam & 61.61 & 68.70 & 64.54 & \cellcolor{gray!50}\textbf{71.31}  &  65.83\\

harbor & 60.05 & 60.40 & 60.47 & \cellcolor{gray!50}\textbf{62.03} &  46.79\\

bridge & 30.48 & 36.74 & 34.24 & 37.94 & \cellcolor{gray!50}\textbf{55.56} \\

vehicle & 42.60 & 47.63 & 43.12 & 49.66 & \cellcolor{gray!50}\textbf{55.10} \\

windmill & 35.32 & 41.99 & 40.76 & 46.72 & \cellcolor{gray!50}\textbf{62.06} \\

airplane & 55.29& 60.17 & 56.43& 60.32&  \cellcolor{gray!50}\textbf{72.95} \\
\midrule
average & 62.44 & 65.13 & 64.12 & 66.46 &  \cellcolor{gray!50}\textbf{72.06} \\
\bottomrule
\end{tabular}
}
\vspace{-0.0cm}
\end{table*}

\subsubsection{Quantitative Results on the RISBench Dataset}

RISBench further extends its dataset scale by incorporating 26 common ground object categories. This expansion includes a broader range of image resolutions and increases the number of attribute categories to eight, posing greater challenges for referring image segmentation tasks. As detailed in Tab. \ref{tab:comparisons-RISBench}, the overall experimental results indicate that RSRefSeg 2 achieves optimal performance across nearly all evaluation metrics, demonstrating significant improvement over existing state-of-the-art methods. RSRefSeg 2 is shown to comprehensively surpass comparative approaches in precision across different threshold intervals. Notably, this method exhibits distinct advantages under higher threshold conditions, enhancing accuracy in fine-grained segmentation. While achieving performance comparable to the recently proposed LSCF method on the cIoU metric, RSRefSeg 2 substantially outperforms it on the gIoU metric.

\begin{table*}[!htbp]
\centering
\caption{Performance comparison across various evaluation metrics on the RISBench test dataset.} \label{tab:comparisons-RISBench}
\vspace{-0.2cm}
\resizebox{0.95\linewidth}{!}{
\begin{tabular}{ c c | *{5}{c} | c c}
\toprule
\textbf{Method} & \textbf{Publication} & \textbf{Pr@0.5} & \textbf{Pr@0.6} & \textbf{Pr@0.7} & \textbf{Pr@0.8} & \textbf{Pr@0.9} & \textbf{cIoU} & \textbf{gIoU} \\ 
\midrule
RRN \cite{li2018referring} & CVPR'18 & 55.04 & 47.31 & 39.86 & 32.58 & 13.24 & 49.67 & 43.18 \\
MAttNet \cite{yu2018mattnet} & CVPR'18 & 56.83 & 48.02 & 41.75 & 34.18 & 15.26 & 51.24 & 45.71 \\
BRINet \cite{hu2020bi} & CVPR'20 & 52.87 & 45.39 & 38.64 & 30.79 & 11.86 & 48.73 & 42.91 \\
LSCM \cite{hui2020linguistic} & ECCV'20 & 55.26 & 47.14 & 40.10 & 33.29 & 13.91 & 50.08 & 43.69 \\
CMPC \cite{huang2020referring} & CVPR'20 & 55.17 & 47.84 & 40.28 & 32.87 & 14.55 & 50.24 & 43.82 \\
CMPC+ \cite{liu2021cross} & TPAMI'21 & 58.02 & 49.00 & 42.53 & 35.26 & 17.88 & 53.98 & 46.73 \\
CRIS \cite{wang2022cris} & CVPR'22 & 63.67 & 55.73 & 44.42 & 28.80 & 13.27 & 69.11 & 55.18 \\
LAVT \cite{yang2022lavt} & CVPR'22 & 69.40 & 63.66 & 56.10 & 44.95 & 25.21 & 74.15 & 61.93 \\
ETRIS \cite{xu2023bridging}& ICCV'23& 60.98 & 51.88 & 39.87 & 24.49 & 11.18 & 67.61 & 53.06 \\
CrossVLT \cite{cho2023cross} & TMM'23 & 70.62 & 65.05 & 57.40 & 45.80 & 26.10 & 74.33 & 62.84 \\
RIS-DMMI \cite{hu2023beyond} & ICCV'23 & 72.05 & 66.48 & 59.07 & 47.16 & 26.57 & 74.82 & 63.93 \\
CARIS \cite{liu2023caris} & ACM MM'23 & 73.94 & 68.93 & 62.08 & 50.31 & 29.08 & 75.10 & 65.79 \\
robust-ref-seg \cite{wu2024towards}& TIP'24  &  69.15 & 63.24 & 55.33 & 43.27 & 24.20 & 74.23 & 61.25 \\
LGCE  \cite{yuan2024rrsis} & TGRS'24 & 69.64 & 64.07 & 56.26 & 44.92 & 25.74 & 73.87 & 62.13 \\
RMSIN \cite{liu2024rotated} & CVPR'24 & 71.01 & 65.46 & 57.69 & 45.50 & 25.92 & 74.09 & 63.07 \\
CroBIM-Swin \cite{dong2024cross} & Arxiv'24 & 75.75 & 70.34 & 63.12 & 51.12 & 28.45 & 73.61 & 67.32 \\
CroBIM-ConvNeXt \cite{dong2024cross} & Arxiv'24 & 77.55 & 72.83 & 66.38 & 55.93 & 34.07 & 73.04 & 69.33 \\
LSCF \cite{ma2025lscf} & TGRS'25 & 76.08 & 71.29 & 64.96 & 55.13  &36.73  &\cellcolor{gray!50}\textbf{74.88} & 68.53 \\
\midrule
RSRefSeg 2 (Ours)  & - & \cellcolor{gray!50}\textbf{79.08} & \cellcolor{gray!50}\textbf{75.36} & \cellcolor{gray!50}\textbf{70.62} & \cellcolor{gray!50}\textbf{62.75} & \cellcolor{gray!50}\textbf{48.75} & 74.77 & \cellcolor{gray!50}\textbf{72.57} \\
\bottomrule
\end{tabular}
}
\vspace{-0.6cm}
\end{table*}

\subsubsection{Qualitative Visualizations}

\begin{figure*}[!htpb]
\centering
\resizebox{0.99\linewidth}{!}{
\includegraphics[width=\linewidth]{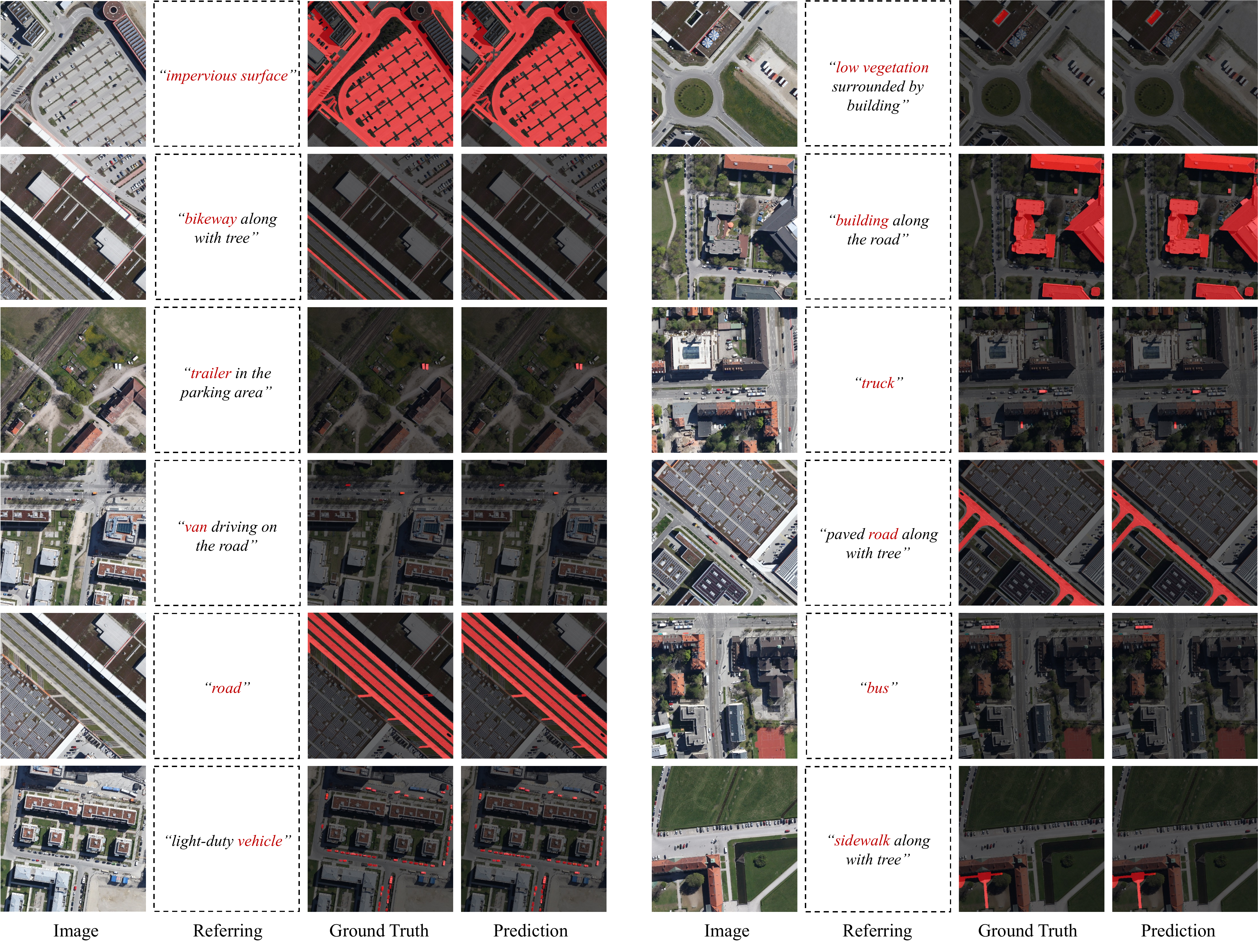}
}
\vspace{-0.2cm}
\caption{
Segmentation examples of the proposed RSRefSeg 2 on the RefSegRS test dataset.
}
\label{fig:vis_RefSegRS}
\vspace{-0.6cm}
\end{figure*}

\begin{figure*}[!htpb]
\centering
\resizebox{0.99\linewidth}{!}{
\includegraphics[width=\linewidth]{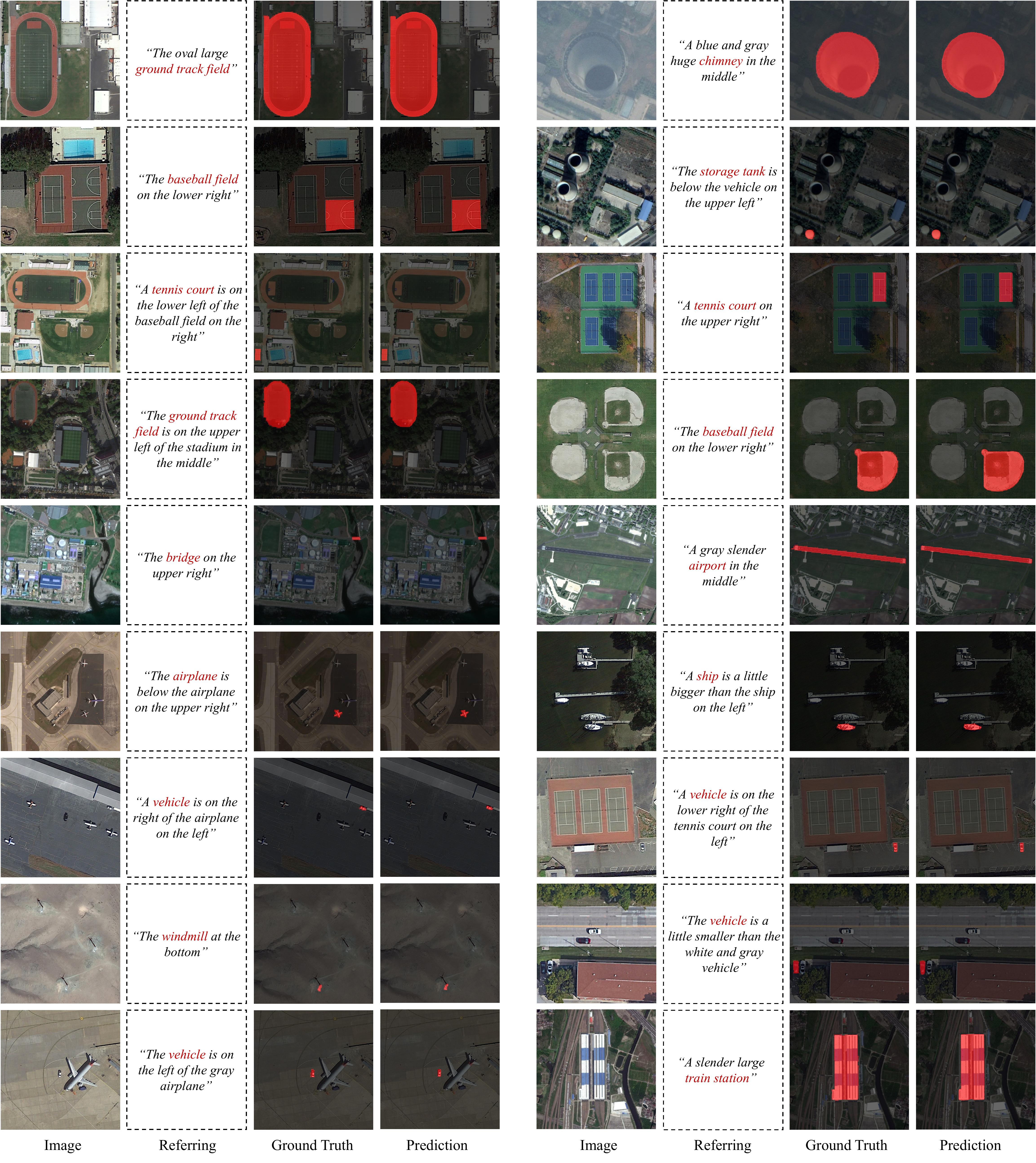}
}
\caption{
Segmentation examples of the proposed RSRefSeg 2 on the RRSIS-D test dataset.
}
\label{fig:vis_RRSIS-D}
\end{figure*}

\begin{figure*}[!htpb]
\centering
\resizebox{0.99\linewidth}{!}{
\includegraphics[width=\linewidth]{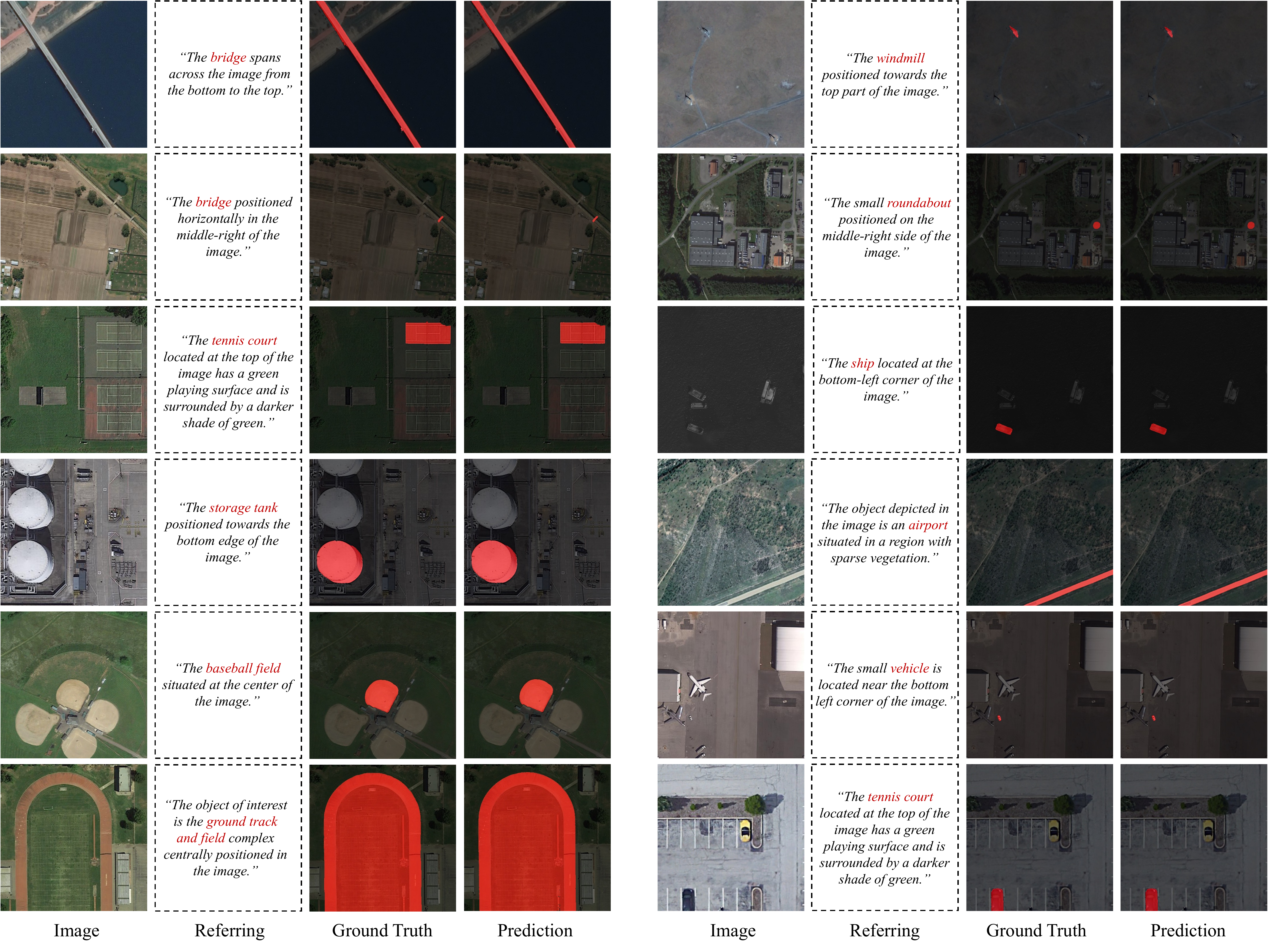}
}
\vspace{-0.2cm}
\caption{
Segmentation examples of the proposed RSRefSeg 2 on the RISBench test dataset.
}
\label{fig:vis_RISBench}
\vspace{-0.6cm}
\end{figure*}

As illustrated in Fig. \ref{fig:vis_RefSegRS}, Fig. \ref{fig:vis_RRSIS-D}, and Fig. \ref{fig:vis_RISBench}, representative segmentation instances from each dataset are presented, demonstrating the correspondence between input images, textual descriptions, ground truth masks, and model predictions. The model exhibits the following capabilities: i) complex referring instructions, such as spatial descriptions (\eg, ``at the top right") and color attributes, are effectively interpreted to achieve precise target segmentation; ii) diverse object categories are accurately segmented, including sports venues (\eg, track fields, baseball fields), infrastructure (\eg, storage tanks, bridges, airports), and vehicles (\eg, cars, buses); iii) high-quality segmentation masks are produced, exhibiting strong alignment with ground truth annotations, characterized by clear edge contours and complete coverage; iv) multi-scale processing is supported, enabling the segmentation of large-scale targets (\eg, track fields, airports) as well as fine-grained small targets (\eg, vehicles); v) excellent detail discrimination ability is demonstrated, characterized by fine-grained perception and semantic understanding, particularly for tasks requiring the distinction between similar objects (\eg, ``The aircraft to the left of the gray plane" or ``The ship slightly larger than the ship on the left").

\subsection{Ablation Study}

This section performs an ablation study on the RRSIS-D dataset to systematically evaluate the contributions of individual components and parameter configurations within the proposed framework. Unless otherwise specified, all experiments utilize referring descriptions encoded by SigLip \cite{tschannen2025siglip} as semantic prompts, with referring mask extraction being performed by SAM2.1-s \cite{ravi2024sam}.

\subsubsection{Effects of Tuning Parameters in RSRefSeg 2 Encoders}

RSRefSeg 2 leverages CLIP's multimodal understanding to achieve coarse localization of textual descriptions within images and generate semantic prompts, subsequently employing SAM's general segmentation capability to refine precise masks. As billion-parameter vision foundation models, CLIP and SAM exhibit exceptional generalization and versatility. However, directly training all parameters risks severe overfitting and performance deterioration, while fully freezing them exacerbates the effects of domain shift. To mitigate this, a low-rank adaptation (LoRA) strategy is adopted, selectively modifying a subset of parameters within the Transformer layers of each encoder. The impact of varying LoRA ranks on the CLIP text encoder (clip-t), CLIP vision encoder (clip-v), and SAM vision encoder (sam-v) is analyzed, as illustrated in Tab. \ref{tab:ab_various_tuning_parameters}. A rank of 0 signifies fully frozen parameters; the optimal configuration ($^\dag$) utilizes ranks of 16, 16, and 32 for clip-t, clip-v, and sam-v, respectively. For each encoder analyzed in the table sections, only its rank is varied while others remain fixed at these optimal values.

Key observations include: i) Parameter fine-tuning significantly enhances performance, yielding greater gains in the vision encoders (clip-v and sam-v) than in the text encoder (clip-t). This discrepancy arises because both vision encoders, pretrained on natural images, confront substantial domain shift in remote sensing contexts, whereas linguistic descriptions exhibit minimal divergence between natural and remote sensing domains. ii) Balancing trainable parameters is critical; insufficient adaptation impedes cross-domain knowledge transfer, while excessive parameters induce overfitting despite accelerated training convergence. iii) Vision encoders require higher ranks than the text encoder, reflecting not only the greater domain shift in visual modalities but also the increased complexity of remote sensing imagery, necessitating more adaptable parameters. iv) Performance is more sensitive to rank variations in the CLIP vision encoder than in SAM. This underscores the pivotal role of semantic prompt precision, as SAM’s segmentation generalizability remains robust even under limited parameter adjustments. Finally, a rank of 32 is utilized for the SAM visual encoder in the SAM2.1-s version, while a rank of 16 is employed for the SAM2.1-l version.

\begin{table}[!htbp] 
\centering
\caption{
Performance with varying LoRA ranks for clip-t, clip-v, and sam-v encoders. 
}
\label{tab:ab_various_tuning_parameters}
\vspace{-0.2cm}
\resizebox{1\linewidth}{!}{
\begin{tabular}{*{2}{c} | *{3}{c} | *{2}{c}}
\toprule
\textbf{Enc.}  & \textbf{Rank} & \textbf{Pr@0.5} & \textbf{Pr@0.7} & \textbf{Pr@0.9} & \textbf{cIoU} & \textbf{gIoU} 
\\ 
\midrule
\multirow{6}{*}{clip-t} &
0 & 72.16 & 55.93 & 27.92 & 77.80 & 63.40 
\\
&4 & 73.48 & 57.94 & 28.69 & 78.65 & 64.61 
\\
&8 & 74.20 &  57.79  & 28.41 &  78.44 &  64.82
\\
&16$^\dag$ & 76.60 & 60.25 & 28.58 & 78.15 & 66.01
\\
&32 & 73.63 & 57.94 & 29.50 & 79.01 & 64.53
\\
&64 & 72.76 & 57.16 & 28.35 & 78.43 & 64.14
\\
\midrule
\multirow{6}{*}{clip-v} &
0 & 65.48 & 50.90 & 26.68 & 75.96 & 57.86 
\\
&4 & 73.42 & 58.20 & 29.81 & 78.44 & 64.64
\\
&8 & 73.83 & 57.22 & 29.18 & 78.64 & 64.33
\\
&16$^\dag$ & 76.60 & 60.25 & 28.58 & 78.15 & 66.01
\\
&32 & 75.23 & 58.46 & 29.07 & 79.36 & 65.39
\\
&64 & 73.97 & 57.85 & 29.30 & 78.96 & 64.88
\\
\midrule
\multirow{6}{*}{sam-v} &
0 & 69.52 & 52.31 & 24.82 & 76.91 & 60.82 
\\
&4 & 73.57 & 57.88 & 29.15 & 78.60 & 64.64 
\\
&8 & 73.71 & 57.82 & 28.78 & 78.38 & 64.67 
\\
&16 & 74.46 & 58.23 & 28.87 & 78.62 & 65.05
\\
&32$^\dag$ & 76.60 & 60.25 & 28.58 & 78.15 & 66.01
\\
&64 & 75.00 & 58.23 & 29.27 & 78.98 & 65.44
\\
\bottomrule
\end{tabular}
}
\vspace{-0.8cm}
\end{table}

\subsubsection{Effects of Varying Module Depths on the Cascaded Referring Prompter}

The prompter consists of three sequential modules: referring semantic decomposition, text-visual interaction, and sparse-dense prompt generation. Each module incorporates several serially-connected blocks, with the number of blocks denoted as $N_\text{decomp}$, $N_\text{inter-act}$, and $N_\text{p-gen}$, respectively (Sec. \ref{sec:prompter}). The impact of varying these block counts is presented in Tab. \ref{tab:ab_various_number_layers}, where the best performance is marked with $^\dag$. Experimental observations indicate that employing two blocks for all three modules achieves an optimal trade-off between model complexity and performance. An insufficient number of blocks constrains representational capability, whereas an excessive number elevates computational overhead and increases the risk of overfitting.

\begin{table}[!tbp] 
\centering
\caption{
Performance of varying block counts across modules in the cascaded second-order referring prompter.
}
\label{tab:ab_various_number_layers}
\vspace{-0.2cm}
\resizebox{1\linewidth}{!}{
\begin{tabular}{*{2}{c} | *{3}{c} | *{2}{c}}
\toprule
\textbf{Module}  & \textbf{Blocks} & \textbf{Pr@0.5} & \textbf{Pr@0.7} & \textbf{Pr@0.9} & \textbf{cIoU} & \textbf{gIoU} 
\\ 
\midrule
\multirow{4}{*}{Decomp.} &
1 & 73.77 & 58.11 & 29.07 & 78.36 & 64.72
\\
&2$^\dag$ & 76.60 & 60.25 & 28.58 & 78.15 & 66.01
\\
&4 & 74.43 & 58.46 & 29.47 & 78.91 & 65.20
\\
&6 & 74.46 & 58.32 & 28.81 & 78.73 & 65.14
\\
\midrule
\multirow{4}{*}{Interac.} &
1 & 74.46 & 58.97 & 29.10 & 78.51 & 64.86
\\
&2$^\dag$ & 76.60 & 60.25 & 28.58 & 78.15 & 66.01
\\
&4 & 74.08 & 57.71 & 28.95 & 78.77 & 64.65
\\
&6 & 73.31 & 57.42 & 28.52 & 78.06 & 64.29
\\
\midrule
\multirow{4}{*}{Prompt} &
1 & 74.40 & 58.51 & 29.18 & 78.95 & 65.04
\\
&2$^\dag$ & 76.60 & 60.25 & 28.58 & 78.15 & 66.01
\\
&4 & 74.08 & 58.03 & 29.44 & 78.43 & 65.00
\\
&6 & 74.60 & 58.54 & 29.13 & 78.49 & 64.88
\\
\bottomrule
\end{tabular}
}
\vspace{-0.25cm}
\end{table}

\subsubsection{Effects of Embedding Vector Count in Semantic Decomposition and Prompt Generation}

Query embeddings are introduced to represent semantic subspaces during decomposition and sparse prompts during prompt generation. These embeddings aggregate textual semantic features and prompt positional information, respectively. The model's expressive capacity is positively correlated with the number of embeddings, denoted as $n_t$ for textual queries and $n_p$ for prompt queries (Sec. \ref{sec:prompter}). As shown in Tab. \ref{tab:ab_various_number_queries_prompts}, optimal performance for the decomposition module (Decomp.) is achieved when representing distinct semantic subspaces with 3 embeddings. For the prompt generation module (Prompt), utilizing 9 embeddings to represent sparse prompts yields optimal performance. It should be noted that optimal settings are dataset-dependent; empirical evidence indicates increasing $n_t$ and $n_p$ enhances expressiveness for longer, complex textual descriptions or images with higher instance counts, whereas reducing them may suffice for simpler cases. Although these parameters are dataset-specific, we adopt unified configurations of $n_t=3$ and $n_p=9$ without exhaustive per-dataset tuning, which nevertheless achieves exceptional performance across all three datasets. This outcome comprehensively outperforms existing comparative methods and underscores the proposed architecture's strong robustness across diverse scenarios.

\begin{table}[!tbp] 
\centering
\caption{Performance of varying numbers of embedding vectors in semantic decomposition and prompt generation. Optimal configurations are marked with $^\dag$.
}
\label{tab:ab_various_number_queries_prompts}
\vspace{-0.2cm}
\resizebox{1\linewidth}{!}{
\begin{tabular}{*{2}{c} | *{3}{c} | *{2}{c}}
\toprule
\textbf{Module}  & \textbf{Embeddings} & \textbf{Pr@0.5} & \textbf{Pr@0.7} & \textbf{Pr@0.9} & \textbf{cIoU} & \textbf{gIoU} 
\\ 
\midrule
\multirow{4}{*}{Decomp.} &
1 & 74.66 & 58.46 & 28.87 & 78.28 & 65.26
\\
&3$^\dag$ & 76.60 & 60.25 & 28.58 & 78.15 & 66.01
\\
&6 & 75.15 & 58.48 & 29.38 & 78.96 & 65.18
\\
&9 & 74.66 & 56.88 & 29.38 & 77.96 & 64.09
\\
\midrule
\multirow{5}{*}{Prompt} &
1 & 73.59 & 57.59 & 29.21 & 78.66 & 64.49
\\
&3 & 74.37 & 57.65 & 28.21 & 78.30 & 64.51
\\
&6 & 73.71 & 57.88 & 29.10 & 78.33 & 64.72
\\
&9$^\dag$ & 76.60 & 60.25 & 28.58 & 78.15 & 66.01
\\
&12 & 74.46 & 57.82 & 29.10 & 79.11 & 65.22
\\
\bottomrule
\end{tabular}
}
\vspace{-0.6cm}
\end{table}

\subsubsection{Effects of Various Component Configurations in the Prompter}

This section analyzes the effects of incorporating positional encoding and various embedding methods. Notation includes `\vmark' for inclusion of positional encoding, `\xmark' for its absence, `zero' for embeddings initialized as zero vectors, `learnable' for embeddings initialized as randomly sampled learnable vectors, `textpool' for embeddings derived from pooled original text features, and `sine' for fixed sine/cosine coordinate-based encoding vectors. Tab. \ref{tab:ab_various_queries_prompts_structures} reveals several key findings: i) Positional encoding is critical across all modules; its inclusion (`\vmark') consistently yields significant performance improvements. Notably, substantial performance gains are observed upon adding positional encoding within the decomposition module, irrespective of the embedding method employed. ii) For the referring semantic decomposition module, the highest performance is achieved using `textpool' embeddings combined with positional encoding, followed by `zero' initialized embeddings with positional encoding; learnable vectors were found to be suboptimal in this context. iii) Within the text-visual interaction module, optimal effectiveness for positional encoding applied to visual feature maps is attained using `learnable' embeddings; conversely, `sine' positional encoding demonstrates no significant advantage. iv) In the prompt generation module, optimal performance is achieved using `learnable' embedding combined with positional encoding.

\begin{table}[!tbp] 
\centering
\caption{Performance of positional encoding and embedding methods across prompter modules.
}
\label{tab:ab_various_queries_prompts_structures}
\vspace{-0.2cm}
\resizebox{1\linewidth}{!}{
\begin{tabular}{*{3}{c} | *{3}{c} | *{2}{c}}
\toprule
\textbf{Module}  & \textbf{PE} & \textbf{Emb.} & \textbf{Pr@0.5} & \textbf{Pr@0.7} & \textbf{Pr@0.9} & \textbf{cIoU} & \textbf{gIoU} 
\\ 
\midrule
\multirow{5}{*}{Decomp.} 
& \xmark & learnable & 73.02 & 57.25 & 28.00 & 77.72 & 64.09
\\
& \xmark & textpool & 74.03 & 58.26 & 28.55 & 78.27 & 64.82
\\
& \vmark & zero & 74.86 & 58.28 & 29.38 & 79.34 & 65.57
\\
& \vmark & learnable & 74.08 & 58.08 & 28.98 & 78.27 & 64.86
\\
& \vmark & textpool & 76.60 & 60.25 & 28.58 & 78.15 & 66.01
\\
\midrule
\multirow{3}{*}{Interac.} 
& \xmark & - & 74.57 & 58.11 & 28.46 & 77.94 & 64.80
\\
& \vmark & sine & 74.26 & 58.23 & 28.98 & 77.90 & 64.66
\\
& \vmark & learnable &  76.60 & 60.25 & 28.58 & 78.15 & 66.01
\\
\midrule
\multirow{3}{*}{Prompt} 
& \xmark & learnable & 74.54 & 58.14 & 29.13 & 79.11 & 64.98
\\
& \vmark & zero & 75.15 & 57.68 & 28.98 & 78.85 & 65.12
\\
& \vmark & learnable &  76.60 & 60.25 & 28.58 & 78.15 & 66.01
\\
\bottomrule
\end{tabular}
}
\vspace{-0.2cm}
\end{table}

\subsubsection{Effects of Cascaded Referring Prompter}

The core mechanism of the cascaded second-order referring prompter utilizes prompt generation driven by the interaction between two cascaded semantic subspaces and visual features. Tab. \ref{tab:ab_prompter_structures} suggests the following observations: i) The introduction of the cascaded second-order referring prompter significantly enhances all evaluation metrics, demonstrating its efficacy in improving segmentation accuracy and regional localization capability relative to a standard single-stage prompter. ii) Integrating dense prompts within the cascaded prompter architecture enhances prompt generation, yielding additional performance gains; this indicates that dense prompts can further augment model performance. In summary, the experimental results validate the effectiveness of the proposed cascaded second-order referring prompter and highlight the significance of integrating dense prompts with sparse prompts.

\begin{table}[!tbp] 
\centering
\caption{Performance of the cascaded prompter architecture and dense prompting.
}
\label{tab:ab_prompter_structures}
\vspace{-0.2cm}
\resizebox{1\linewidth}{!}{
\begin{tabular}{*{2}{c} | *{3}{c} | *{2}{c}}
\toprule
\textbf{Cascaded Prompter} & \textbf{Dense Prompt} &  \textbf{Pr@0.5} & \textbf{Pr@0.7} & \textbf{Pr@0.9} & \textbf{cIoU} & \textbf{gIoU} 
\\ 
\midrule
 \xmark & \xmark & 67.88 & 52.57 & 25.39 & 74.11 & 60.06
\\
 \vmark & \xmark & 72.99 & 57.39 & 28.12 & 79.22 & 64.13
\\
 \vmark & \vmark &  76.60 & 60.25 & 28.58 & 78.15 & 66.01
\\
\bottomrule
\end{tabular}
}
\vspace{-0.6cm}
\end{table}

\subsubsection{Effects of Subspace Attention Mechanism in Referring Semantic Decomposition}

Referring semantic decomposition facilitates implicit cascaded reasoning by decomposing the original referring text into dual semantic subspaces through three principal mechanisms: an intra-subspace self-attention mechanism operating within each semantic subspace, a cross-attention mechanism bridging the subspaces and the original text embeddings, and an external cross-attention mechanism enabling interaction between the subspaces (Eq. \ref{eq:decomposition}). Tab. \ref{tab:ab_query_to_text_structures} demonstrates that optimal performance is achieved when intra-subspace self-attention is combined with inter-subspace cross-attention (final row). This underscores the necessity of bidirectional exchange between subspaces for mutual comprehension, thereby promoting effective semantic fusion. Furthermore, the intra-subspace self-attention mechanism is essential for establishing accurate semantic representations within each subspace and enabling subspace-specific modeling.

\begin{table}[!tbp] 
\centering
\caption{Performance of intra-subspace self-attention and inter-subspace cross-attention.
}
\label{tab:ab_query_to_text_structures}
\vspace{-0.2cm}
\resizebox{1\linewidth}{!}{
\begin{tabular}{*{2}{c} | *{3}{c} | *{2}{c}}
\toprule
\textbf{intra-attn}  & \textbf{inter-attn} & \textbf{Pr@0.5} & \textbf{Pr@0.7} & \textbf{Pr@0.9} & \textbf{cIoU} & \textbf{gIoU} 
\\ 
\midrule
 \xmark & \xmark & 72.68 & 57.74 & 29.08 & 78.16 & 64.08
\\
\vmark & \xmark & 74.23 & 58.20 & 29.09 & 78.27 & 64.86
\\
\vmark & \vmark &  76.60 & 60.25 & 28.58 & 78.15 & 66.01
\\
\bottomrule
\end{tabular}
}
\vspace{-0.2cm}
\end{table}

\subsubsection{Effects of Attention Mechanism in Text-Visual Interaction}

Text-visual interaction employs two core attention mechanisms: self-attention (self-attn), which processes features within each semantic subspace, and cross-attention (cross-attn), which facilitates interaction between each subspace and the semantically enhanced visual features. As shown in Table \ref{tab:ab_query_to_img_structures}, cross-attention yielded the most significant performance gains, indicating its effectiveness in enhancing vision-language integration and strengthening the expression of referring expression features. Conversely, employing self-attention alone only marginally improves accuracy.

\begin{table}[!tbp] 
\centering
\caption{Performance of different attention mechanism pairings in the text–visual interaction.
}
\label{tab:ab_query_to_img_structures}
\vspace{-0.2cm}
\resizebox{1\linewidth}{!}{
\begin{tabular}{*{2}{c} | *{3}{c} | *{2}{c}}
\toprule
\textbf{self-attn}  & \textbf{cross-attn} & \textbf{Pr@0.5} & \textbf{Pr@0.7} & \textbf{Pr@0.9} & \textbf{cIoU} & \textbf{gIoU} 
\\ 
\midrule
 \xmark & \xmark & 73.54 & 56.27 & 28.08 & 77.82 & 64.39
\\
\vmark & \xmark & 73.22 & 57.42 & 28.28 & 78.03 & 64.54
\\
\vmark & \vmark &  76.60 & 60.25 & 28.58 & 78.15 & 66.01
\\
\bottomrule
\end{tabular}
}
\vspace{-0.6cm}
\end{table}

\subsubsection{Effects of Dense Prompt Construction Methods}

Dense and sparse prompts jointly provide target location information for the second-stage fine-grained segmentation. As formalized in Eq. \ref{eq:prompt-dense}, dense prompts utilize text embedding vectors as channel-wise filters to weight visual features. Tab. \ref{tab:ab_prompter_structures} verifies the effectiveness of dense prompts, which motivated this comprehensive study evaluating their compositional variants through the experiments summarized in Tab. \ref{tab:ab_dense_promt_structures}.

i) Four distinct text embedding sources were evaluated: a) T-$t_2$, derived from the pooled second semantic subspace vector $t_2$ (Eq. \ref{eq:decomposition}); b) T-$t_2^\prime$, obtained from the pooled vector $t_2^\prime$ following visual-text interaction (Eq. \ref{eq:Interaction}); c) T-sparse, representing the dedicated additional token filter in sparse prompts (Eq. \ref{eq:prompt-sparse}); and d) T-$t_\text{sent}$, the original global semantic vector $t_\text{sent}$ (Sec. \ref{sec:dual-encoder}). Consequently, the original global semantic vector ($t_\text{sent}$) achieved superior performance, significantly outperforming the alternative embeddings. This result validates the importance of utilizing shallow global text semantics for generating effective dense prompts.

ii) The impact of incorporating an MLP layer into text embeddings was evaluated. Performance degradation was observed when MLP processing (w/ T-soft) was implemented, indicating that such additional transformations may disrupt the original semantic representations.

iii) Two visual feature sources were assessed: a) V-$v$, corresponding to the original CLIP features $v$ (Sec. \ref{sec:dual-encoder}); and b) V-$v_2^\prime$, denoting the features $v_2^\prime$ after text-guided interaction (Eq. \ref{eq:Interaction}). Superior performance was observed for V-$v_2^\prime$, particularly on comprehensive evaluation metrics, demonstrating that visual-text interaction effectively refines referring features, thereby improving segmentation performance.

iv) The effect of applying a $1\times1$ convolutional layer to visual features was examined. A moderate performance improvement was observed when visual features were processed (w/ V-soft), suggesting that such adaptive soft adjustment effectively enhances visual representations, contrasting with text embedding conclusions.

v) The upsampling strategy was explored since CLIP feature dimensions ($h_1 < \frac{H_2}{4}$, $w_1 < \frac{W_2}{4}$) are insufficient for SAM's dense prompt requirements, necessitating bilinear interpolation. Two approaches were compared: a) Pre-up, interpolation applied to visual features before filtering; b) Post-up, interpolation applied after filtering. Pre-filtering interpolation (Pre-up) achieved better performance, indicating that early-stage upsampling better preserves fine-grained information.

In summary, the optimal dense prompt configuration employs: the original global text semantic vector ($t_\text{sent}$) without MLP processing; post-interaction visual features ($v_2^\prime$) with convolutional processing; and pre-filtering upsampling. This design harmonizes textual and visual fusion mechanisms to provide accurate spatial guidance for fine-grained segmentation.

\begin{table}[!tbp] 
\centering
\caption{
Performance of varying dense prompt construction methods.
}
\label{tab:ab_dense_promt_structures}
\vspace{-0.2cm}
\resizebox{0.96\linewidth}{!}{
\begin{tabular}{c | *{3}{c} | *{2}{c}}
\toprule
\textbf{Setting}  & \textbf{Pr@0.5} & \textbf{Pr@0.7} & \textbf{Pr@0.9} & \textbf{cIoU} & \textbf{gIoU} 
\\ 
\midrule

T-$t_2$&  73.80 & 58.05 & 28.69 & 77.79 & 64.44
\\
T-$t_2^\prime$ & 71.07 & 55.55 & 28.21 & 76.33 & 62.58
\\
T-sparse & 71.10 & 56.42 & 27.92 & 76.60 & 63.08
\\
T-$t_\text{sent}$ &  76.60 & 60.25 & 28.58 & 78.15 & 66.01
\\
\midrule
w/ T-soft & 74.51 & 57.65 & 28.84 & 77.86 & 64.74
\\
w/o T-soft & 76.60 & 60.25 & 28.58 & 78.15 & 66.01
\\
\midrule
V-$v$ & 73.80 & 57.82 & 28.92 & 78.67 & 64.93
\\
V-$v_2^\prime$ & 76.60 & 60.25 & 28.58 & 78.15 & 66.01
\\
\midrule
w/o V-soft & 74.11 & 59.20 & 29.55 & 78.58 & 65.02
\\
w/ V-soft &  76.60 & 60.25 & 28.58 & 78.15 & 66.01
\\
\midrule
post-up & 74.66 & 59.83 & 28.89 & 77.47 & 65.21
\\
pre-up & 76.60 & 60.25 & 28.58 & 78.15 & 66.01
\\
\bottomrule
\end{tabular}
}
\vspace{-0.2cm}
\end{table}

\begin{table}[!t] 
\centering
\caption{Performance of varying input resolutions of CLIP visual encoder.
}
\label{tab:ab_clip_size}
\vspace{-0.2cm}
\resizebox{0.8\linewidth}{!}{
\begin{tabular}{*{1}{c} | *{3}{c} | *{2}{c}}
\toprule
\textbf{Size} & \textbf{Pr@0.5} & \textbf{Pr@0.7} & \textbf{Pr@0.9} & \textbf{cIoU} & \textbf{gIoU} 
\\ 
\midrule
224$^2$ & 76.47 & 60.21 & 28.52 & 78.06 & 65.85 
\\
384$^2$ & 76.60 & 60.25 & 28.58 & 78.15 & 66.01
\\
448$^2$ &  76.64 & 60.07 & 28.69 & 78.57 & 66.10
\\
768$^2$ &  76.68 & 60.29 & 29.15 & 78.76 & 66.14
\\
\bottomrule
\end{tabular}
}
\vspace{-0.6cm}
\end{table}

\subsubsection{Effects of Input Size on CLIP Visual Encoder}

Tab. \ref{tab:ab_clip_size} presents the model performance across varying input sizes. The results demonstrate a consistent improvement in segmentation accuracy as the input resolution increases, indicating that higher-resolution inputs contribute positively to performance. Nonetheless, the observed gains are relatively modest, supporting the hypothesis that coarse localization tasks can be effectively performed even at lower resolutions. Considering the trade-off between computational efficiency and segmentation performance, an input size of $384 \times 384$ was selected for the small model variant.

\subsubsection{Analysis of Loss Function Components}

As detailed in Sec. \ref{sec:loss}, the overall loss function integrates several key components. Systematic ablation studies of these components (Tab. \ref{tab:ab_varying_loss}) demonstrate their distinct contributions to model performance. Under cross-entropy segmentation supervision $\mathcal{L}_{\text{ce}}$, a substantial enhancement in metrics is observed upon introduction of the semantic subspace constraint $\mathcal{L}_{\text{ortho}}$, indicating its efficacy in improving segmentation accuracy. Subsequent integration of dense prompt supervision $\mathcal{L}_{\text{dense}}$, however, yields negligible gains. This limitation is potentially attributed to the direct supervision of low-resolution coarse segmentation masks, where loss of fine details may restrict improvements. Multiple implementations of the spatial constraint $\mathcal{L}_{\text{spatial}}$ (\eg, CE-$t_{\text{sent}}$-$v$, MIL-$t_{\text{sent}}$-$v$) were evaluated. Both cross-entropy (CE) and Multi-Instance Learning (MIL) approaches yielded marginal improvements, suggesting constraints applied to the spatial dimension of visual feature maps offer limited efficacy. Conversely, sample-wise NCE learning applied to original CLIP visual features generated substantial performance gains, with the NCE-$t_2$-$v$ configuration demonstrating notable superiority. Ultimately, integration of the segmentation Dice loss $\mathcal{L}_{\text{dice}}$ further enhances model performance. In summary, this ablation study reveals that: semantic subspace constraints improve segmentation accuracy; supervision based on low-resolution dense prompts proves suboptimal; sample-wise NCE learning on visual features surpasses purely spatial constraints; and Dice loss integration significantly strengthens segmentation performance.

\begin{table}[!tbp] 
\centering
\caption{
Effects of different loss function components.
}
\label{tab:ab_varying_loss}
\vspace{-0.2cm}
\resizebox{1\linewidth}{!}{
\begin{tabular}{*{6}{c} | *{3}{c} | *{2}{c}}
\toprule
$\mathcal{L}_{\text{ce}}$ & $\mathcal{L}_{\text{ortho}}$  & $\mathcal{L}_{\text{dense}}$ & $\mathcal{L}_{\text{spatial}}$ & $\mathcal{L}_{\text{sample}}$ & $\mathcal{L}_{\text{dice}}$ & \textbf{Pr@0.5} & \textbf{Pr@0.7} & \textbf{Pr@0.9} & \textbf{cIoU} & \textbf{gIoU} 
\\ 
\midrule
\vmark & - & - &  -&-  &-  & 79.72 & 55.41 & 28.03 & 78.32 & 63.16
\\
\vmark & \vmark & - & - & - &  -& 74.62 & 57.28 & 28.58 & 78.37 & 64.21
\\
\vmark & \vmark& \vmark & - & - &- & 73.13 & 57.82 & 29.21 & 78.91 & 64.08
\\
\midrule
\vmark & \vmark & - & CE-$t_\text{sent}$-$v$ &- & -& 72.47 & 58.11 & 28.98 & 77.03 & 63.12
\\
\vmark & \vmark & -& MIL-$t_\text{sent}$-$v$ &- &- & 71.21 & 57.36 & 28.98 & 78.27 & 63.01
\\
\vmark & \vmark &- & CE-$t_2$-$v$ & -&- & 72.56 & 58.02 & 29.30 & 77.54 & 63.65
\\
\vmark & \vmark & -& MIL-$t_2$-$v$ &- & -& 73.54 & 58.26 & 28.44 & 78.80 & 64.25
\\
\vmark & \vmark &- & CE-$t_\text{sent}$-$v_2^\prime$ &- &- & 72.65 & 57.51 & 29.50 & 76.88 & 63.77
\\
\vmark & \vmark & - & MIL-$t_\text{sent}$-$v_2^\prime$ &- & -& 73.71 & 57.65 & 28.95 & 77.93 & 64.02
\\
\vmark & \vmark & -& CE-$t_2$-$v_2^\prime$ &- & -& 72.96 & 58.31 & 29.53 & 77.83 & 63.95
\\
\vmark & \vmark &- & MIL-$t_2$-$v_2^\prime$ & -& -& 74.11 & 59.00 & 29.53 & 78.52 & 64.65
\\
\midrule
\vmark & \vmark & -&-  & NCE-$t_\text{sent}$-$v$ &-& 74.40 & 58.69 & 29.24 & 78.78 & 65.18
\\
\vmark & \vmark & -&- & NCE-$t_2$-$v$ &  -&76.60 & 60.25 & 28.58 & 78.15 & 66.01
\\
\vmark & \vmark &- &- & NCE-$t_\text{sent}$-$v_2^\prime$ &-& 71.90 & 56.62 & 28.46 & 77.32 & 63.13
\\
\vmark & \vmark &- & - & NCE-$t_2$-$v_2^\prime$ &-& 72.65 & 57.25 & 28.09 & 77.16 & 63.52
\\
\midrule
\vmark & \vmark &- &- & NCE-$t_2$-$v$ & \vmark& 77.90 & 61.13 & 29.35 & 78.98 & 67.06
\\
\bottomrule
\end{tabular}
}
\vspace{-0.6cm}
\end{table}

\subsubsection{Effects of Different Backbone Versions}

Tab. \ref{tab:ab_backbone_structures} presents various combinations of CLIP and SAM model versions and their influence on performance. First, under a fixed SAM configuration (sam2.1-s), increasing the CLIP model’s input resolution yields significantly greater performance improvements compared to augmenting its parameter count. This indicates that higher input resolution enhances target localization accuracy, a characteristic particularly advantageous for remote sensing scenarios characterized by small, widely distributed targets. Conversely, with a fixed CLIP model (siglip2-p16-512), segmentation performance progressively improves as SAM model parameters increase, achieving optimal results with sam2.1-l. These results confirm the pivotal role of foundation model capacity during the refined segmentation stage in boosting accuracy. Consequently, variants of differing specifications may be selected according to computational resource constraints in practical applications to optimize the performance-to-resource consumption ratio.

\vspace{-0.4cm}
\subsection{Discussions}

\begin{table}[!tbp] 
\centering
\caption{
Performance comparison of various CLIP and SAM model combinations.
}
\label{tab:ab_backbone_structures}
\vspace{-0.1cm}
\resizebox{1\linewidth}{!}{
\begin{tabular}{*{2}{c} | *{3}{c} | *{2}{c}}
\toprule
\textbf{CLIP}  & \textbf{SAM} & \textbf{Pr@0.5} & \textbf{Pr@0.7} & \textbf{Pr@0.9} & \textbf{cIoU} & \textbf{gIoU} 
\\ 
\midrule
siglip-p14-384 & sam2.1-s & 76.70 & 60.35 & 28.58 & 78.05 & 66.13
\\
siglip2-p14-224 & sam2.1-s & 75.58 & 59.32 & 28.26 & 77.55 & 65.36
\\
\midrule
siglip2-p14-224 & sam2.1-s & 75.58 & 59.32 & 28.26 & 77.55 & 65.36
\\
siglip2-p14-384 & sam2.1-s & 77.90 & 61.13 & 29.35 & 78.98 & 67.06
\\
siglip2-p16-384 & sam2.1-s & 77.33 & 60.72 & 28.66 & 78.29 & 66.73
\\
siglip2-p16-512 & sam2.1-s & 78.94 & 61.79 & 29.07 & 78.43 & 67.79
\\
siglip2-giant-p16-256 & sam2.1-s & 76.29 & 60.27 & 28.32 & 77.17 & 65.62
\\
siglip2-giant-p16-384 & sam2.1-s & 77.67 & 60.87 & 28.82 & 77.52 & 66.94
\\
\midrule
siglip2-p16-512 & sam2.1-t & 78.56 & 61.56 & 27.83 & 78.48 & 67.63
\\
siglip2-p16-512 & sam2.1-s & 78.94 & 61.79 & 29.07 & 78.43 & 67.79
\\
siglip2-p16-512 & sam2.1-b & 79.28 & 63.91 & 30.45 & 78.52 & 68.07
\\
siglip2-p16-512 & sam2.1-l & 80.23 & 65.41 & 31.05 & 79.45 & 69.17
\\
\bottomrule
\end{tabular}
}
\vspace{-0.2cm}
\end{table}

\begin{figure}[!tpb]
\centering
\resizebox{\linewidth}{!}{
\includegraphics[width=\linewidth]{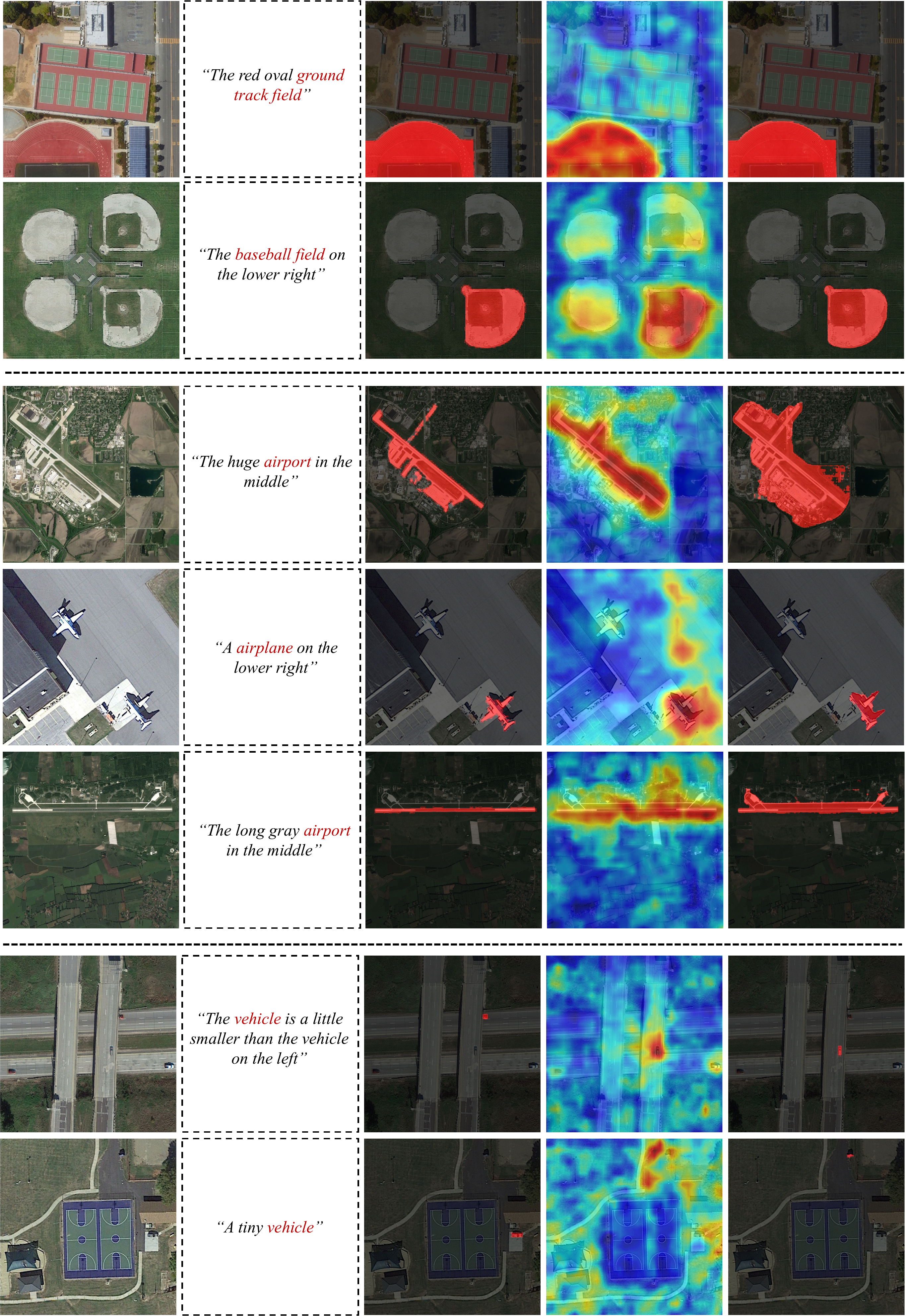}
}
\vspace{-0.2cm}
\caption{
Segmentation results across various cases. From left to right: input image, text instruction, ground truth, dense prompt, and model prediction.
}
\label{fig:fig_discussion}
\vspace{-0.6cm}
\end{figure}

RSRefSeg 2 decomposes the remote sensing referring segmentation task into two sequential stages: ``coarse localization" and ``fine-grained segmentation." This approach employs a cascaded two-stage prompt mechanism to guide a pre-trained multi-modal encoder (CLIP), activating text-guided visual features that subsequently drive a segmentation foundation model (SAM) to produce the final mask. Visualization analyses indicate that the model resolves complex spatial directional instructions (\eg, ``upper right," ``lower left") and multi-attribute descriptions accurately. It demonstrates superior performance in multi-scale object segmentation, diverse category recognition, and boundary clarity, particularly excelling at segmenting small targets (\eg, bridges, vehicles). This advantage originates from the dual-gain mechanism of the decoupled architecture: during the localization stage, orthogonal subspace projection within the text embedding space effectively mitigates target misactivation caused by multi-entity descriptions; during the segmentation stage, task decoupling isolates the mask generation process from text modality interference, thereby ensuring masks exhibit sharp edges and intact structures.

Fig. \ref{fig:fig_discussion} visualizes segmentation results across various cases. The figure displays, from left to right: the input image, text instruction, ground truth, dense prompt, and model prediction. In the first row, both the dense prompt and final prediction are effective, achieving high segmentation accuracy. The second row illustrates accurate localization but suboptimal segmentation deviating from the ground truth, reflecting limitations in the fine-grained segmentation model under specific conditions. The third row exhibits a localization error despite a plausible segmentation result, indicating limitations in the text-visual alignment capability of the localization stage. The current decoupled architecture facilitates clear identification of performance bottlenecks. The separability analysis of errors, attributing semantic comprehension failures to the localization stage and boundary fuzziness to the segmentation module, provides clear direction for subsequent iterative optimization.

However, limitations persist in the following scenarios: First, for descriptions containing compound semantics (\eg, ``a bright airplane located at the bottom edge of the image with its wings extending to the right"), fine-grained text-visual alignment discrepancies remain, primarily due to the current cross-modal model's limited capability for modeling complex logical relationships. Second, in low-contrast scenarios or when processing objects with indistinct boundaries (\eg, dams, harbor outlines), SAM's mask generation mechanism is susceptible to erroneous predictions. This issue is directly linked to SAM's prior knowledge, predominantly based on clear boundaries in natural images. Future research could explore two optimization avenues: on one hand, introducing reinforcement learning frameworks to enhance semantic reasoning capabilities and construct a closed-loop feedback mechanism between language understanding and visual localization; on the other hand, exploring alternative segmentation mechanisms to SAM and developing novel segmentation components adapted to the boundary ambiguity inherent in remote sensing imagery.

\vspace{-0.2cm}
\section{Conclusion}

To address challenges in referring remote sensing image segmentation, specifically multimodal feature misalignment, insufficient small target recognition accuracy, and limited model interpretability, this paper proposes RSRefSeg 2. RSRefSeg 2 incorporates prior knowledge derived from visual foundation models and implements a decoupled segmentation process consisting of two sequential stages: coarse localization and refined segmentation. Leveraging a cascaded second-order prompting mechanism, the approach activates text-guided visual features within the pretrained multimodal encoder CLIP and subsequently directs the foundation segmentation model SAM to generate corresponding masks. During the coarse localization stage, an orthogonality-constrained projection of text embeddings is employed to mitigate target misactivation caused by conflicting multi-entity descriptions, thereby enhancing spatial localization precision. Within the refined segmentation stage, task decoupling ensures mask generation proceeds independently of textual comprehension, yielding masks characterized by well-defined boundaries and structural integrity. Concurrently, parameter-efficient fine-tuning is achieved through the integration of LoRA, enabling effective adaptation to domain-specific remote sensing knowledge while preserving the intrinsic general representation capabilities of the pretrained models. Experimental results (on RefSegRS, RRSIS-D, and RISBench) demonstrate that RSRefSeg 2 achieves significant performance gains, exhibiting particular strength in open-scene semantic understanding, multi-scale object segmentation, small target recognition (\eg, ships and vehicles), and the parsing of complex spatial relationships. Furthermore, by explicitly distinguishing between localization and segmentation errors, model interpretability is enhanced, providing reliable support for advancing remote sensing intelligent interpretation paradigms towards greater efficiency and generalization.

\ifCLASSOPTIONcaptionsoff
  \newpage
\fi



{\scriptsize
\bibliographystyle{IEEEtran}
\bibliography{IEEEabrv,myreferences}

\begin{thebibliography}{10}
\providecommand{\url}[1]{#1}
\csname url@samestyle\endcsname
\providecommand{\newblock}{\relax}
\providecommand{\bibinfo}[2]{#2}
\providecommand{\BIBentrySTDinterwordspacing}{\spaceskip=0pt\relax}
\providecommand{\BIBentryALTinterwordstretchfactor}{4}
\providecommand{\BIBentryALTinterwordspacing}{\spaceskip=\fontdimen2\font plus
\BIBentryALTinterwordstretchfactor\fontdimen3\font minus \fontdimen4\font\relax}
\providecommand{\BIBforeignlanguage}[2]{{%
\expandafter\ifx\csname l@#1\endcsname\relax
\typeout{** WARNING: IEEEtran.bst: No hyphenation pattern has been}%
\typeout{** loaded for the language `#1'. Using the pattern for}%
\typeout{** the default language instead.}%
\else
\language=\csname l@#1\endcsname
\fi
#2}}
\providecommand{\BIBdecl}{\relax}
\BIBdecl

\bibitem{yuan2024rrsis}
Z.~Yuan, L.~Mou, Y.~Hua, and X.~X. Zhu, ``Rrsis: Referring remote sensing image segmentation,'' \emph{IEEE Transactions on Geoscience and Remote Sensing}, 2024.

\bibitem{chen2025rsrefseg}
K.~Chen, J.~Zhang, C.~Liu, Z.~Zou, and Z.~Shi, ``Rsrefseg: Referring remote sensing image segmentation with foundation models,'' \emph{arXiv preprint arXiv:2501.06809}, 2025.

\bibitem{dong2024cross}
Z.~Dong, Y.~Sun, T.~Liu, W.~Zuo, and Y.~Gu, ``Cross-modal bidirectional interaction model for referring remote sensing image segmentation,'' \emph{arXiv preprint arXiv:2410.08613}, 2024.

\bibitem{dong2025diffris}
Z.~Dong, Y.~Sun, T.~Liu, and Y.~Gu, ``Diffris: Enhancing referring remote sensing image segmentation with pre-trained text-to-image diffusion models,'' \emph{arXiv preprint arXiv:2506.18946}, 2025.

\bibitem{liu2024rotated}
S.~Liu, Y.~Ma, X.~Zhang, H.~Wang, J.~Ji, X.~Sun, and R.~Ji, ``Rotated multi-scale interaction network for referring remote sensing image segmentation,'' in \emph{Proceedings of the IEEE/CVF Conference on Computer Vision and Pattern Recognition}, 2024, pp. 26\,658--26\,668.

\bibitem{lei2024exploring}
S.~Lei, X.~Xiao, T.~Zhang, H.-C. Li, Z.~Shi, and Q.~Zhu, ``Exploring fine-grained image-text alignment for referring remote sensing image segmentation,'' \emph{IEEE Transactions on Geoscience and Remote Sensing}, 2024.

\bibitem{chen2021building}
K.~Chen, Z.~Zou, and Z.~Shi, ``Building extraction from remote sensing images with sparse token transformers,'' \emph{Remote Sensing}, vol.~13, no.~21, p. 4441, 2021.

\bibitem{yang2025large}
Z.~Yang, H.~Yao, L.~Tian, X.~Zhao, Q.~Li, and Q.~Wang, ``A large-scale referring remote sensing image segmentation dataset and benchmark,'' \emph{arXiv preprint arXiv:2506.03583}, 2025.

\bibitem{ma2025lscf}
Q.~Ma, L.~Li, X.~Lu, L.~Jiao, F.~Liu, W.~Ma, X.~Liu, and L.~Sun, ``Lscf: Long-term semantic-guidance convformer for referring remote sensing image segmentation,'' \emph{IEEE Transactions on Geoscience and Remote Sensing}, 2025.

\bibitem{liu2022dual}
Z.~Liu, K.~Hao, X.~Geng, Z.~Zou, and Z.~Shi, ``Dual-branched spatio-temporal fusion network for multihorizon tropical cyclone track forecast,'' \emph{IEEE Journal of Selected Topics in Applied Earth Observations and Remote Sensing}, vol.~15, pp. 3842--3852, 2022.

\bibitem{liu2024deriving}
Z.~Liu, H.~Chen, L.~Bai, W.~Li, K.~Chen, Z.~Wang, W.~Ouyang, Z.~Zou, and Z.~Shi, ``Deriving accurate surface meteorological states at arbitrary locations via observation-guided continous neural field modeling,'' \emph{IEEE Transactions on Geoscience and Remote Sensing}, 2024.

\bibitem{chen2023ovarnet}
K.~Chen, X.~Jiang, Y.~Hu, X.~Tang, Y.~Gao, J.~Chen, and W.~Xie, ``Ovarnet: Towards open-vocabulary object attribute recognition,'' in \emph{Proceedings of the IEEE/CVF conference on computer vision and pattern recognition}, 2023, pp. 23\,518--23\,527.

\bibitem{zou2023object}
Z.~Zou, K.~Chen, Z.~Shi, Y.~Guo, and J.~Ye, ``Object detection in 20 years: A survey,'' \emph{Proceedings of the IEEE}, vol. 111, no.~3, pp. 257--276, 2023.

\bibitem{zhang2025referring}
T.~Zhang, Z.~Wen, B.~Kong, K.~Liu, Y.~Zhang, P.~Zhuang, and J.~Li, ``Referring remote sensing image segmentation via bidirectional alignment guided joint prediction,'' \emph{arXiv preprint arXiv:2502.08486}, 2025.

\bibitem{ho2024rssep}
N.-V. Ho, T.~Phan, M.~Adkins, C.~Rainwater, J.~Cothren, and N.~Le, ``Rssep: Sequence-to-sequence model for simultaneous referring remote sensing segmentation and detection,'' in \emph{Proceedings of the Asian Conference on Computer Vision}, 2024, pp. 218--231.

\bibitem{pan2024rethinking}
Y.~Pan, R.~Sun, Y.~Wang, T.~Zhang, and Y.~Zhang, ``Rethinking the implicit optimization paradigm with dual alignments for referring remote sensing image segmentation,'' in \emph{Proceedings of the 32nd ACM International Conference on Multimedia}, 2024, pp. 2031--2040.

\bibitem{li2025multimodal}
Y.~Li, W.~Jin, S.~Qiu, and Q.~Sun, ``Multimodal prompt-guided bidirectional fusion for referring remote sensing image segmentation,'' \emph{Remote Sensing}, vol.~17, no.~10, p. 1683, 2025.

\bibitem{liu2025cadformer}
M.~Liu, X.~Jiang, and X.~Zhang, ``Cadformer: Fine-grained cross-modal alignment and decoding transformer for referring remote sensing image segmentation,'' \emph{IEEE Journal of Selected Topics in Applied Earth Observations and Remote Sensing}, 2025.

\bibitem{radford2021learning}
A.~Radford, J.~W. Kim, C.~Hallacy, A.~Ramesh, G.~Goh, S.~Agarwal, G.~Sastry, A.~Askell, P.~Mishkin, J.~Clark \emph{et~al.}, ``Learning transferable visual models from natural language supervision,'' in \emph{International conference on machine learning}.\hskip 1em plus 0.5em minus 0.4em\relax PmLR, 2021, pp. 8748--8763.

\bibitem{chen2025dynamicvis}
K.~Chen, C.~Liu, B.~Chen, W.~Li, Z.~Zou, and Z.~Shi, ``Dynamicvis: An efficient and general visual foundation model for remote sensing image understanding,'' \emph{arXiv preprint arXiv:2503.16426}, 2025.

\bibitem{shi2025multimodal}
L.~Shi and J.~Zhang, ``Multimodal-aware fusion network for referring remote sensing image segmentation,'' \emph{IEEE Geoscience and Remote Sensing Letters}, 2025.

\bibitem{li2025semantic}
S.~Li, S.~Wang, Z.~Sun, and J.~Xiao, ``Semantic localization guiding segment anything model for reference remote sensing image segmentation,'' \emph{arXiv preprint arXiv:2506.10503}, 2025.

\bibitem{li2025scale}
K.~Li, G.~Vosselman, and M.~Y. Yang, ``Scale-wise bidirectional alignment network for referring remote sensing image segmentation,'' \emph{arXiv preprint arXiv:2501.00851}, 2025.

\bibitem{li2025aeroreformer}
R.~Li and X.~Zhao, ``Aeroreformer: Aerial referring transformer for uav-based referring image segmentation,'' \emph{arXiv preprint arXiv:2502.16680}, 2025.

\bibitem{kirillov2023segment}
A.~Kirillov, E.~Mintun, N.~Ravi, H.~Mao, C.~Rolland, L.~Gustafson, T.~Xiao, S.~Whitehead, A.~C. Berg, W.-Y. Lo \emph{et~al.}, ``Segment anything,'' in \emph{Proceedings of the IEEE/CVF international conference on computer vision}, 2023, pp. 4015--4026.

\bibitem{liu2024change}
C.~Liu, K.~Chen, H.~Zhang, Z.~Qi, Z.~Zou, and Z.~Shi, ``Change-agent: Towards interactive comprehensive remote sensing change interpretation and analysis,'' \emph{IEEE Transactions on Geoscience and Remote Sensing}, 2024.

\bibitem{chen2022resolution}
K.~Chen, W.~Li, J.~Chen, Z.~Zou, and Z.~Shi, ``Resolution-agnostic remote sensing scene classification with implicit neural representations,'' \emph{IEEE Geoscience and Remote Sensing Letters}, vol.~20, pp. 1--5, 2022.

\bibitem{liu2025text2earth}
C.~Liu, K.~Chen, R.~Zhao, Z.~Zou, and Z.~Shi, ``Text2earth: Unlocking text-driven remote sensing image generation with a global-scale dataset and a foundation model,'' \emph{arXiv preprint arXiv:2501.00895}, 2025.

\bibitem{chen2024rsmamba}
K.~Chen, B.~Chen, C.~Liu, W.~Li, Z.~Zou, and Z.~Shi, ``Rsmamba: Remote sensing image classification with state space model,'' \emph{IEEE Geoscience and Remote Sensing Letters}, 2024.

\bibitem{li2021geographical}
W.~Li, K.~Chen, H.~Chen, and Z.~Shi, ``Geographical knowledge-driven representation learning for remote sensing images,'' \emph{IEEE Transactions on Geoscience and Remote Sensing}, vol.~60, pp. 1--16, 2021.

\bibitem{li2025segearth}
K.~Li, Z.~Xin, L.~Pang, C.~Pang, Y.~Deng, J.~Yao, G.~Xia, D.~Meng, Z.~Wang, and X.~Cao, ``Segearth-r1: Geospatial pixel reasoning via large language model,'' \emph{arXiv preprint arXiv:2504.09644}, 2025.

\bibitem{yang2022lavt}
Z.~Yang, J.~Wang, Y.~Tang, K.~Chen, H.~Zhao, and P.~H. Torr, ``Lavt: Language-aware vision transformer for referring image segmentation,'' in \emph{Proceedings of the IEEE/CVF conference on computer vision and pattern recognition}, 2022, pp. 18\,155--18\,165.

\bibitem{pan2025mixed}
X.~Pan, X.~Xie, and J.~Yang, ``Mixed-scale cross-modal fusion network for referring image segmentation,'' \emph{Neurocomputing}, vol. 614, p. 128793, 2025.

\bibitem{li2022geographical}
W.~Li, K.~Chen, and Z.~Shi, ``Geographical supervision correction for remote sensing representation learning,'' \emph{IEEE Transactions on Geoscience and Remote Sensing}, vol.~60, pp. 1--20, 2022.

\bibitem{chen2024time}
K.~Chen, C.~Liu, W.~Li, Z.~Liu, H.~Chen, H.~Zhang, Z.~Zou, and Z.~Shi, ``Time travelling pixels: Bitemporal features integration with foundation model for remote sensing image change detection,'' in \emph{IGARSS 2024-2024 IEEE International Geoscience and Remote Sensing Symposium}.\hskip 1em plus 0.5em minus 0.4em\relax IEEE, 2024, pp. 8581--8584.

\bibitem{liu2024rscama}
C.~Liu, K.~Chen, B.~Chen, H.~Zhang, Z.~Zou, and Z.~Shi, ``Rscama: Remote sensing image change captioning with state space model,'' \emph{IEEE Geoscience and Remote Sensing Letters}, 2024.

\bibitem{chen2023continuous}
K.~Chen, W.~Li, S.~Lei, J.~Chen, X.~Jiang, Z.~Zou, and Z.~Shi, ``Continuous remote sensing image super-resolution based on context interaction in implicit function space,'' \emph{IEEE Transactions on Geoscience and Remote Sensing}, vol.~61, pp. 1--16, 2023.

\bibitem{chen2025seg}
B.~Chen, K.~Chen, M.~Yang, Z.~Zou, and Z.~Shi, ``Seg-sr: Integrating semantic knowledge into remote sensing image super-resolution via vision-language model,'' \emph{arXiv preprint arXiv:2505.23010}, 2025.

\bibitem{bai2023qwen}
J.~Bai, S.~Bai, Y.~Chu, Z.~Cui, K.~Dang, X.~Deng, Y.~Fan, W.~Ge, Y.~Han, F.~Huang \emph{et~al.}, ``Qwen technical report,'' \emph{arXiv preprint arXiv:2309.16609}, 2023.

\bibitem{rombach2022high}
R.~Rombach, A.~Blattmann, D.~Lorenz, P.~Esser, and B.~Ommer, ``High-resolution image synthesis with latent diffusion models,'' in \emph{Proceedings of the IEEE/CVF conference on computer vision and pattern recognition}, 2022, pp. 10\,684--10\,695.

\bibitem{ravi2024sam}
N.~Ravi, V.~Gabeur, Y.-T. Hu, R.~Hu, C.~Ryali, T.~Ma, H.~Khedr, R.~R{\"a}dle, C.~Rolland, L.~Gustafson \emph{et~al.}, ``Sam 2: Segment anything in images and videos,'' \emph{arXiv preprint arXiv:2408.00714}, 2024.

\bibitem{chen2024rsprompter}
K.~Chen, C.~Liu, H.~Chen, H.~Zhang, W.~Li, Z.~Zou, and Z.~Shi, ``Rsprompter: Learning to prompt for remote sensing instance segmentation based on visual foundation model,'' \emph{IEEE Transactions on Geoscience and Remote Sensing}, vol.~62, pp. 1--17, 2024.

\bibitem{liu2024remoteclip}
F.~Liu, D.~Chen, Z.~Guan, X.~Zhou, J.~Zhu, Q.~Ye, L.~Fu, and J.~Zhou, ``Remoteclip: A vision language foundation model for remote sensing,'' \emph{IEEE Transactions on Geoscience and Remote Sensing}, 2024.

\bibitem{guo2024skysense}
X.~Guo, J.~Lao, B.~Dang, Y.~Zhang, L.~Yu, L.~Ru, L.~Zhong, Z.~Huang, K.~Wu, D.~Hu \emph{et~al.}, ``Skysense: A multi-modal remote sensing foundation model towards universal interpretation for earth observation imagery,'' in \emph{Proceedings of the IEEE/CVF Conference on Computer Vision and Pattern Recognition}, 2024, pp. 27\,672--27\,683.

\bibitem{wang2024mtp}
D.~Wang, J.~Zhang, M.~Xu, L.~Liu, D.~Wang, E.~Gao, C.~Han, H.~Guo, B.~Du, D.~Tao \emph{et~al.}, ``Mtp: Advancing remote sensing foundation model via multi-task pretraining,'' \emph{IEEE Journal of Selected Topics in Applied Earth Observations and Remote Sensing}, 2024.

\bibitem{li2025agrifm}
W.~Li, S.~Liang, K.~Chen, Y.~Chen, H.~Ma, J.~Xu, Y.~Ma, S.~Guan, H.~Fang, and Z.~Shi, ``Agrifm: A multi-source temporal remote sensing foundation model for crop mapping,'' \emph{arXiv preprint arXiv:2505.21357}, 2025.

\bibitem{sun2022ringmo}
X.~Sun, P.~Wang, W.~Lu, Z.~Zhu, X.~Lu, Q.~He, J.~Li, X.~Rong, Z.~Yang, H.~Chang \emph{et~al.}, ``Ringmo: A remote sensing foundation model with masked image modeling,'' \emph{IEEE Transactions on Geoscience and Remote Sensing}, vol.~61, pp. 1--22, 2022.

\bibitem{bi2025ringmoe}
H.~Bi, Y.~Feng, B.~Tong, M.~Wang, H.~Yu, Y.~Mao, H.~Chang, W.~Diao, P.~Wang, Y.~Yu \emph{et~al.}, ``Ringmoe: Mixture-of-modality-experts multi-modal foundation models for universal remote sensing image interpretation,'' \emph{arXiv preprint arXiv:2504.03166}, 2025.

\bibitem{hong2024spectralgpt}
D.~Hong, B.~Zhang, X.~Li, Y.~Li, C.~Li, J.~Yao, N.~Yokoya, H.~Li, P.~Ghamisi, X.~Jia \emph{et~al.}, ``Spectralgpt: Spectral remote sensing foundation model,'' \emph{IEEE Transactions on Pattern Analysis and Machine Intelligence}, 2024.

\bibitem{wang2025hypersigma}
D.~Wang, M.~Hu, Y.~Jin, Y.~Miao, J.~Yang, Y.~Xu, X.~Qin, J.~Ma, L.~Sun, C.~Li \emph{et~al.}, ``Hypersigma: Hyperspectral intelligence comprehension foundation model,'' \emph{IEEE Transactions on Pattern Analysis and Machine Intelligence}, 2025.

\bibitem{hu2025diffusion}
Y.~Hu, L.~Zhang, X.~Luo, and X.~Cao, ``Diffusion self-distillation for remote sensing scene classification,'' \emph{IEEE Transactions on Geoscience and Remote Sensing}, 2025.

\bibitem{zheng2024segment}
Z.~Zheng, Y.~Zhong, L.~Zhang, and S.~Ermon, ``Segment any change,'' \emph{arXiv preprint arXiv:2402.01188}, 2024.

\bibitem{gong2024crossearth}
Z.~Gong, Z.~Wei, D.~Wang, X.~Ma, H.~Chen, Y.~Jia, Y.~Deng, Z.~Ji, X.~Zhu, N.~Yokoya \emph{et~al.}, ``Crossearth: Geospatial vision foundation model for domain generalizable remote sensing semantic segmentation,'' \emph{arXiv preprint arXiv:2410.22629}, 2024.

\bibitem{zhang2022dino}
H.~Zhang, F.~Li, S.~Liu, L.~Zhang, H.~Su, J.~Zhu, L.~M. Ni, and H.-Y. Shum, ``Dino: Detr with improved denoising anchor boxes for end-to-end object detection,'' \emph{arXiv preprint arXiv:2203.03605}, 2022.

\bibitem{li2022blip}
J.~Li, D.~Li, C.~Xiong, and S.~Hoi, ``Blip: Bootstrapping language-image pre-training for unified vision-language understanding and generation,'' in \emph{International conference on machine learning}.\hskip 1em plus 0.5em minus 0.4em\relax PMLR, 2022, pp. 12\,888--12\,900.

\bibitem{zhou2022learning}
K.~Zhou, J.~Yang, C.~C. Loy, and Z.~Liu, ``Learning to prompt for vision-language models,'' \emph{International Journal of Computer Vision}, vol. 130, no.~9, pp. 2337--2348, 2022.

\bibitem{ding2021openprompt}
N.~Ding, S.~Hu, W.~Zhao, Y.~Chen, Z.~Liu, H.-T. Zheng, and M.~Sun, ``Openprompt: An open-source framework for prompt-learning,'' \emph{arXiv preprint arXiv:2111.01998}, 2021.

\bibitem{khattak2023maple}
M.~U. Khattak, H.~Rasheed, M.~Maaz, S.~Khan, and F.~S. Khan, ``Maple: Multi-modal prompt learning,'' in \emph{Proceedings of the IEEE/CVF conference on computer vision and pattern recognition}, 2023, pp. 19\,113--19\,122.

\bibitem{zhou2022conditional}
K.~Zhou, J.~Yang, C.~C. Loy, and Z.~Liu, ``Conditional prompt learning for vision-language models,'' in \emph{Proceedings of the IEEE/CVF conference on computer vision and pattern recognition}, 2022, pp. 16\,816--16\,825.

\bibitem{lei2024prompt}
Y.~Lei, J.~Li, Z.~Li, Y.~Cao, and H.~Shan, ``Prompt learning in computer vision: a survey,'' \emph{Frontiers of Information Technology \& Electronic Engineering}, vol.~25, no.~1, pp. 42--63, 2024.

\bibitem{han2024parameter}
Z.~Han, C.~Gao, J.~Liu, J.~Zhang, and S.~Q. Zhang, ``Parameter-efficient fine-tuning for large models: A comprehensive survey,'' \emph{arXiv preprint arXiv:2403.14608}, 2024.

\bibitem{xin2024parameter}
Y.~Xin, J.~Yang, S.~Luo, H.~Zhou, J.~Du, X.~Liu, Y.~Fan, Q.~Li, and Y.~Du, ``Parameter-efficient fine-tuning for pre-trained vision models: A survey,'' \emph{arXiv preprint arXiv:2402.02242}, 2024.

\bibitem{jia2022visual}
M.~Jia, L.~Tang, B.-C. Chen, C.~Cardie, S.~Belongie, B.~Hariharan, and S.-N. Lim, ``Visual prompt tuning,'' in \emph{European conference on computer vision}.\hskip 1em plus 0.5em minus 0.4em\relax Springer, 2022, pp. 709--727.

\bibitem{li2023rs}
X.~Li, C.~Wen, Y.~Hu, and N.~Zhou, ``Rs-clip: Zero shot remote sensing scene classification via contrastive vision-language supervision,'' \emph{International Journal of Applied Earth Observation and Geoinformation}, vol. 124, p. 103497, 2023.

\bibitem{zhong2022regionclip}
Y.~Zhong, J.~Yang, P.~Zhang, C.~Li, N.~Codella, L.~H. Li, L.~Zhou, X.~Dai, L.~Yuan, Y.~Li \emph{et~al.}, ``Regionclip: Region-based language-image pretraining,'' in \emph{Proceedings of the IEEE/CVF conference on computer vision and pattern recognition}, 2022, pp. 16\,793--16\,803.

\bibitem{hu2022lora}
E.~J. Hu, Y.~Shen, P.~Wallis, Z.~Allen-Zhu, Y.~Li, S.~Wang, L.~Wang, W.~Chen \emph{et~al.}, ``Lora: Low-rank adaptation of large language models.'' \emph{ICLR}, vol.~1, no.~2, p.~3, 2022.

\bibitem{zhang2025cdmamba}
H.~Zhang, K.~Chen, C.~Liu, H.~Chen, Z.~Zou, and Z.~Shi, ``Cdmamba: Incorporating local clues into mamba for remote sensing image binary change detection,'' \emph{IEEE Transactions on Geoscience and Remote Sensing}, 2025.

\bibitem{azimi2019skyscapes}
S.~M. Azimi, C.~Henry, L.~Sommer, A.~Schumann, and E.~Vig, ``Skyscapes fine-grained semantic understanding of aerial scenes,'' in \emph{Proceedings of the IEEE/CVF International Conference on Computer Vision}, 2019, pp. 7393--7403.

\bibitem{zhan2023rsvg}
Y.~Zhan, Z.~Xiong, and Y.~Yuan, ``Rsvg: Exploring data and models for visual grounding on remote sensing data,'' \emph{IEEE Transactions on Geoscience and Remote Sensing}, vol.~61, pp. 1--13, 2023.

\bibitem{li2024vrsbench}
X.~Li, J.~Ding, and M.~Elhoseiny, ``Vrsbench: A versatile vision-language benchmark dataset for remote sensing image understanding,'' \emph{arXiv preprint arXiv:2406.12384}, 2024.

\bibitem{hu2016segmentation}
R.~Hu, M.~Rohrbach, and T.~Darrell, ``Segmentation from natural language expressions,'' in \emph{Computer Vision--ECCV 2016: 14th European Conference, Amsterdam, The Netherlands, October 11--14, 2016, Proceedings, Part I 14}.\hskip 1em plus 0.5em minus 0.4em\relax Springer, 2016, pp. 108--124.

\bibitem{li2018referring}
R.~Li, K.~Li, Y.-C. Kuo, M.~Shu, X.~Qi, X.~Shen, and J.~Jia, ``Referring image segmentation via recurrent refinement networks,'' in \emph{Proceedings of the IEEE Conference on Computer Vision and Pattern Recognition}, 2018, pp. 5745--5753.

\bibitem{ye2019cross}
L.~Ye, M.~Rochan, Z.~Liu, and Y.~Wang, ``Cross-modal self-attention network for referring image segmentation,'' in \emph{Proceedings of the IEEE/CVF conference on computer vision and pattern recognition}, 2019, pp. 10\,502--10\,511.

\bibitem{hu2020bi}
Z.~Hu, G.~Feng, J.~Sun, L.~Zhang, and H.~Lu, ``Bi-directional relationship inferring network for referring image segmentation,'' in \emph{Proceedings of the IEEE/CVF conference on computer vision and pattern recognition}, 2020, pp. 4424--4433.

\bibitem{hui2020linguistic}
T.~Hui, S.~Liu, S.~Huang, G.~Li, S.~Yu, F.~Zhang, and J.~Han, ``Linguistic structure guided context modeling for referring image segmentation,'' in \emph{European Conference on Computer Vision}.\hskip 1em plus 0.5em minus 0.4em\relax Springer, 2020, pp. 59--75.

\bibitem{huang2020referring}
S.~Huang, T.~Hui, S.~Liu, G.~Li, Y.~Wei, J.~Han, L.~Liu, and B.~Li, ``Referring image segmentation via cross-modal progressive comprehension,'' in \emph{Proceedings of the IEEE/CVF conference on computer vision and pattern recognition}, 2020, pp. 10\,488--10\,497.

\bibitem{liu2021cross}
S.~Liu, T.~Hui, S.~Huang, Y.~Wei, B.~Li, and G.~Li, ``Cross-modal progressive comprehension for referring segmentation,'' \emph{IEEE Transactions on Pattern Analysis and Machine Intelligence}, vol.~44, no.~9, pp. 4761--4775, 2021.

\bibitem{wang2022cris}
Z.~Wang, Y.~Lu, Q.~Li, X.~Tao, Y.~Guo, M.~Gong, and T.~Liu, ``Cris: Clip-driven referring image segmentation,'' in \emph{Proceedings of the IEEE/CVF conference on computer vision and pattern recognition}, 2022, pp. 11\,686--11\,695.

\bibitem{liu2023caris}
S.-A. Liu, Y.~Zhang, Z.~Qiu, H.~Xie, Y.~Zhang, and T.~Yao, ``Caris: Context-aware referring image segmentation,'' in \emph{Proceedings of the 31st ACM International Conference on Multimedia}, 2023, pp. 779--788.

\bibitem{hu2023beyond}
Y.~Hu, Q.~Wang, W.~Shao, E.~Xie, Z.~Li, J.~Han, and P.~Luo, ``Beyond one-to-one: Rethinking the referring image segmentation,'' in \emph{Proceedings of the IEEE/CVF International Conference on Computer Vision}, 2023, pp. 4067--4077.

\bibitem{cho2023cross}
Y.~Cho, H.~Yu, and S.-J. Kang, ``Cross-aware early fusion with stage-divided vision and language transformer encoders for referring image segmentation,'' \emph{IEEE Transactions on Multimedia}, vol.~26, pp. 5823--5833, 2023.

\bibitem{zhang2024evf}
Y.~Zhang, T.~Cheng, L.~Zhu, R.~Hu, L.~Liu, H.~Liu, L.~Ran, X.~Chen, W.~Liu, and X.~Wang, ``Evf-sam: Early vision-language fusion for text-prompted segment anything model,'' \emph{arXiv preprint arXiv:2406.20076}, 2024.

\bibitem{rong2025customized}
F.~Rong, M.~Lan, Q.~Zhang, and L.~Zhang, ``Customized sam 2 for referring remote sensing image segmentation,'' \emph{arXiv preprint arXiv:2503.07266}, 2025.

\bibitem{tschannen2025siglip}
M.~Tschannen, A.~Gritsenko, X.~Wang, M.~F. Naeem, I.~Alabdulmohsin, N.~Parthasarathy, T.~Evans, L.~Beyer, Y.~Xia, B.~Mustafa \emph{et~al.}, ``Siglip 2: Multilingual vision-language encoders with improved semantic understanding, localization, and dense features,'' \emph{arXiv preprint arXiv:2502.14786}, 2025.

\bibitem{yu2018mattnet}
L.~Yu, Z.~Lin, X.~Shen, J.~Yang, X.~Lu, M.~Bansal, and T.~L. Berg, ``Mattnet: Modular attention network for referring expression comprehension,'' in \emph{Proceedings of the IEEE conference on computer vision and pattern recognition}, 2018, pp. 1307--1315.

\bibitem{xu2023bridging}
Z.~Xu, Z.~Chen, Y.~Zhang, Y.~Song, X.~Wan, and G.~Li, ``Bridging vision and language encoders: Parameter-efficient tuning for referring image segmentation,'' in \emph{Proceedings of the IEEE/CVF international conference on computer vision}, 2023, pp. 17\,503--17\,512.

\bibitem{wu2024towards}
J.~Wu, X.~Li, X.~Li, H.~Ding, Y.~Tong, and D.~Tao, ``Towards robust referring image segmentation,'' \emph{IEEE Transactions on Image Processing}, 2024.

\end{thebibliography}
}

\end{document}